\pdfoutput=1

\documentclass[11pt]{article}

\usepackage[preprint]{acl}

\usepackage{times}
\usepackage{latexsym}
\usepackage{graphicx}
\usepackage{subfigure}
\usepackage{booktabs} 

\usepackage{multirow}
\usepackage{verbatim}
\usepackage{graphicx}
\usepackage{pifont}
\usepackage{adjustbox}
\usepackage{enumitem}
\usepackage{amsfonts}
\usepackage{array}
\usepackage{caption}

\usepackage{amssymb}
\usepackage{mathtools}
\usepackage{amsthm}

\usepackage[T1]{fontenc}

\usepackage[utf8]{inputenc}

\usepackage{microtype}

\usepackage{inconsolata}

\usepackage{graphicx}

\title{ChartMind: A Comprehensive Benchmark for \\ Complex Real-world  Multimodal Chart Question Answering}

\author{
  Jingxuan Wei$^{1,2,4}$\thanks{\ \ Equal contribution.},
  Nan Xu$^{1,3}$\footnotemark[1],
  Junnan Zhu$^{3}$\thanks{\ \ Corresponding author.},
  Yanni Hao$^{1,3}$,
  Gaowei Wu$^{2,4}$, \\
  {\bf Qi Chen$^{2,4}$,
  Bihui Yu$^{2,4}$,
  Lei Wang$^{1,3}$} \\
  $^1$Beijing Wenge Technology Co., Ltd. \\
  $^2$Shenyang Institute of Computing Technology, Chinese Academy of Sciences \\
  $^3$MAIS, Institute of Automation, Chinese Academy of Sciences \\
  $^4$University of Chinese Academy of Sciences \\
  \texttt{junnan.zhu@nlpr.ia.ac.cn, weijingxuan20@mails.ucas.edu.cn, xunan2015@ia.ac.cn}
}

\begin{document}
\maketitle
\begin{abstract}
Chart question answering (CQA) has become a critical multimodal task for evaluating the reasoning capabilities of vision-language models. While early approaches have shown promising performance by focusing on visual features or leveraging large-scale pre-training, most existing evaluations rely on rigid output formats and objective metrics, thus ignoring the complex, real-world demands of practical chart analysis. In this paper, we introduce ChartMind, a new benchmark designed for complex CQA tasks in real-world settings. ChartMind covers seven task categories, incorporates multilingual contexts, supports open-domain textual outputs, and accommodates diverse chart formats, bridging the gap between real-world applications and traditional academic benchmarks. Furthermore, we propose a context-aware yet model-agnostic framework, ChartLLM, that focuses on extracting key contextual elements, reducing noise, and enhancing the reasoning accuracy of multimodal large language models. Extensive evaluations on ChartMind and three representative public benchmarks with 14 mainstream multimodal models show our framework significantly outperforms the previous three common CQA paradigms: instruction-following, OCR-enhanced, and chain-of-thought, highlighting the importance of flexible chart understanding for real-world CQA. These findings suggest new directions for developing more robust chart reasoning in future research.
\end{abstract}

\section{Introduction}
\begin{figure*}[t]
    \centering
    \includegraphics[width=\linewidth]{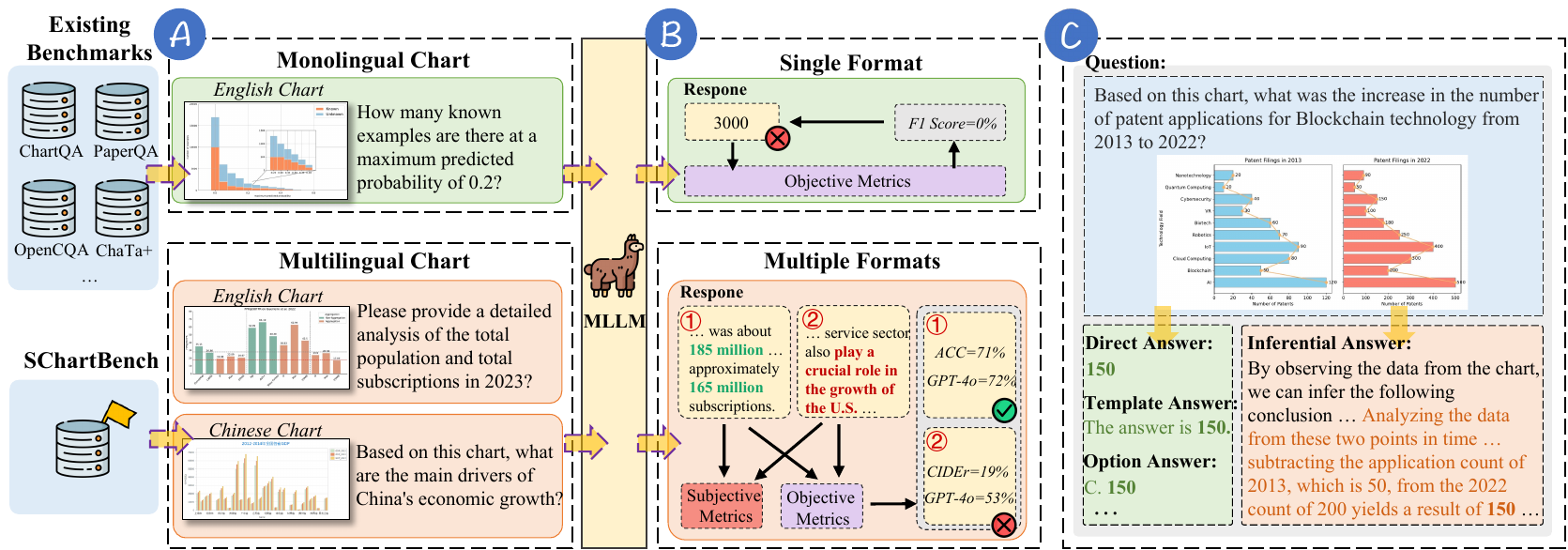}
    \caption{Key Challenges in CQA Benchmarks: (A) Predominantly monolingual, limiting multilingual applicability in chart question answering; (B) Fixed formats and metrics, restricting adaptability to diverse charts; (C) Emphasis on deterministic answers, overlooking complex reasoning, such as  trend analysis, and summarization.}
    \label{fig:challenges}
\end{figure*}


Chart question answering~\cite{ma2024c,qin2022deep} is a prominent multimodal task designed to evaluate the reasoning capabilities of vision-language models, especially their multimodal perception ability and local reasoning ability. Early studies treat CQA as a discriminative task, focusing on directly modeling visual elements to answer questions~\cite{kafle2018dvqa,chang2022mapqa}. However, these methods often struggle with generalization due to their inability to capture the semantic and visual richness of charts. Hence, researchers introduce more visual semantic information (e.g., OCR) to enhance the multimodal perception ability~\cite{liu2023deplot,wang2023domino}. Recent studies have shown the potential of multimodal large language models (LLMs) on the CQA task by adopting large-scale multimodal pre-training~\cite{donut,pix2struct} or chain-of-thought (COT) reasoning~\cite{li2024synthesize,wei2024mchartqa}, suggesting that leveraging large-scale datasets and supervised fine-tuning improves the interpretation of multimodal charts.

Several benchmarks~\cite{zaib2022conversational,bajic2023review,huang2024pixels} have been proposed to better understand the strengths and weaknesses of multi-modal LLMs for CQA. However, human evaluations often suffer from high variability and instability due to individual and cultural differences, leading many existing benchmarks~\cite{kafle2018dvqa,9973197} to rely predominantly on automatic metrics (\textit{e.g.}, F1 scores). While such approaches effectively evaluate the accuracy of a single answer (\textit{e.g.}, ``\textit{2024}" for ``\textit{What is the largest value in column X?}"), they do not fully capture the need for complex and multi-step reasoning commonly required in real-world scenarios. Many professional data analysis tasks demand advanced inference, such as multi-hop reasoning or synthesizing information from multiple charts. Consequently, most existing benchmarks have widely ignored the logical steps involved in such inferencing, focusing instead on whether the answer includes the correct keyword or value. 

In addition, as shown in ~\autoref{fig:challenges}, we summarize three main challenges in existing benchmarks: multilingual charts, diverse formats, and questions lacking a single definitive answer, such as chart summarization. Models need to handle both visual comprehension and logical reasoning. To extract meaningful information, they must first recognize visual elements, such as colors, structures, and spatial relationships. Then, they must analyze the logical connections between elements and answer complex queries, such as performing calculations, identifying trends, and finding relationships within the data. Moreover, the wide range of real-world chart types (\textit{e.g.}, bar charts, line charts, scatter plots) creates higher demands for models to generalize and perform well on new and unseen formats.

To address these challenges, we introduce ChartMind, a multilingual benchmark designed for high-level chart reasoning across seven task categories. It includes both English and Chinese charts, providing the first dual-language evaluation setting for chart QA. Compared to prior benchmarks that focus on single-answer prediction, ChartMind supports open-ended outputs such as summarization and trend analysis. This design narrows the gap between academic benchmarks and real-world chart usage scenarios. To support better performance in these complex tasks, we propose ChartLLM, a structured context modeling framework that explicitly extracts semantic components—titles, legends, axes—from charts and feeds them into the model. Unlike procedural reasoning like CoT, ChartLLM reduces cognitive burden by pre-structuring relevant visual information, improving the robustness and generalizability of existing MLLMs.

To validate our benchmark, we conduct a comprehensive study of 14 mainstream multimodal models, comparing ChartLLM-based approaches with three widely used CQA paradigms: (1) instruction-following methods driven by predefined prompts, (2) OCR-enhanced methods that prioritize text extraction, and (3) COT-based methods emphasizing step-by-step reasoning.

Our contributions are as follows:
(1) We introduce \textbf{ChartMind}, the first benchmark for complex CQA tasks in real-world settings. Covering seven task categories, multilingual contexts, and diverse chart formats, it bridges the gap between real-world applications and traditional academic benchmarks.
(2) We propose \textbf{ChartLLM}, a context-aware yet model-agnostic framework that focuses on extracting key contextual elements, reducing noise, and enhancing the reasoning accuracy of MLLMs.
(3) Through experiments across seven task categories, two languages, and seven chart formats, we show that ChartLLM outperforms prevalent CQA paradigms. 
These findings highlight the need for flexible chart understanding and foster advanced research on real-world chart analysis.

\section{Related Work}

\begin{table*}[ht]
    \centering
    \resizebox{\linewidth}{!}{
    \begin{tabular}{lccccccccccccccc}
        \toprule
        \textbf{Dataset} & \textbf{Avg. Ans.} & \textbf{Instances}  &\textbf{Language} & \textbf{Diverse} & \textbf{Task}   & \textbf{Topic} &  \textbf{Chart} & \textbf{Pie} & \textbf{Scatter} & \textbf{Common} & \textbf{Grouped} & \textbf{Stacked} & \textbf{Complex} & \textbf{Common} \\
        \textbf{} &\textbf{Length} & \textbf{Number} & \textbf{Format} & \textbf{Format} &\textbf{Format}  & \textbf{Format} &  \textbf{Format}  & \textbf{} & \textbf{} & \textbf{Bar} & \textbf{Bar} & \textbf{Bar} & \textbf{Line} & \textbf{Line} \\
        \midrule
        ChartQA~\cite{chartqa} &1.15 & 2,500 & English  & 1 & 1   & 3 &  3& \ding{51} & \ding{55} & \ding{51} & \ding{55} & \ding{55} & \ding{55} & \ding{51} \\
        MMC-Benchmark~\cite{mmc}& 1.08 & 2,126 & English & 1 & 4  & 5 &  2& \ding{55} & \ding{51} & \ding{55} & \ding{55} & \ding{55} & \ding{55} & \ding{51} \\
        PaperQA~\cite{lu2023mathvista} &1.26 & 107 & English  & 1 & 1  & 2 &  4& \ding{51} & \ding{51} & \ding{51} & \ding{55} & \ding{55} & \ding{55} & \ding{51} \\
        OpenCQA~\cite{opencqa}  &55.73 & 1,159 & English & 1 & 1 & 4   &   4&\ding{51} & \ding{51} & \ding{51} & \ding{55} & \ding{55} & \ding{55} & \ding{51} \\
        Chart-to-Text~\cite{charttotext}&73.49 & 3,474 & English  & 1 & 1   & 3 &  4& \ding{51} & \ding{51} & \ding{51} & \ding{55} & \ding{55} & \ding{55} & \ding{51} \\
        LineCap~\cite{9973197}&13.63 & 1,930 & English& 1 & 1   & 1 &  2& \ding{55} & \ding{55} & \ding{55} & \ding{55} & \ding{55} & \ding{51} & \ding{51} \\
        \midrule
        \textbf{ChartMind}& \textbf{119.69} & \textbf{757}&\textbf{EN\&ZH}  & \textbf{2} & \textbf{7}  & \textbf{6}  &  \textbf{7}  & \ding{51} & \ding{51} & \ding{51} & \ding{51} & \ding{51} & \ding{51} & \ding{51} \\
        \bottomrule
    \end{tabular}}
    \caption{Comparison of ChartMind with Existing Chart QA Datasets.}
    \label{tab:dataset_comparison}
     \vskip -0.2in
\end{table*}

In contrast, ChartLLM uses structured semantic cues from charts—such as titles, legends, and axes—to guide model reasoning, without relying on step-by-step decomposition.

\textbf{CQA Methods.} The development of CQA methods~\cite{zeng2024advancing,li2024synthesize,xu2023improving} has evolved from early discriminative approaches to structured reasoning and large-scale pretraining~\cite{zhou2023instruction,li2023improving,huang2024pixels,tan2025boosting}. Early models like IMG+QUESS~\cite{kafle2018dvqa} and V-MODEQA~\cite{chang2022mapqa} use CNNs for visual encoding and RNNs for query processing, but suffer from limited generalization due to weak reasoning and OOV handling. OCR-enhanced methods~\cite{liu2023deplot,wang2023domino} convert chart visuals into text, aiding value extraction but introducing noise and losing spatial cues. COT-based models~\cite{li2024synthesize,wei2024mchartqa} decompose reasoning steps to improve interpretability, yet depend on structured input and struggle with varied chart layouts. Other methods like Donut~\cite{donut} and Pix2Struct~\cite{pix2struct} remove OCR dependency via end-to-end training, while instruction-following models~\cite{gpt4v} leverage large-scale vision-language pretraining but still fall short on multilingual support and high-level reasoning. Recent work such as ChartInsights~\cite{chartinsights} targets low-level factual QA, whereas ChartLLM uses structured semantic cues—titles, legends, axes—to support multilingual and high-level tasks without relying on CoT-style decomposition.

\textbf{CQA Benchmarks.} The development of CQA models necessitates reliable benchmarks to evaluate performance across diverse tasks~\cite{zaib2022conversational,bajic2023review}. Existing datasets fall into Factoid Question Answering (FQA), Open-Domain Question Answering (OQA), and Captioning (CAP) categories~\cite{huang2024pixels}. FQA datasets, such as ChartQA~\cite{kafle2018dvqa}, MMC-Bench~\cite{mmc}, and PaperQA~\cite{lu2023mathvista}, assess factual queries, including numerical extractions, trend identification, and relational interpretations, relying on predefined chart types for objective reasoning. OQA datasets like OpenCQA~\cite{opencqa} introduce open-ended questions but enforce rigid output structures and rely on automated metrics like BLEU, limiting adaptability to complex reasoning. CAP datasets, including Chart-to-Text~\cite{charttotext} and LineCap~\cite{9973197}, generate textual chart descriptions but remain constrained by structured evaluation metrics. ChartMind addresses these gaps by combining high-level semantic tasks, multilingual data, and diverse chart types to support broader and more flexible evaluation. Table \ref{tab:dataset_comparison} compares representative CQA benchmarks.

\section{Construction of ChartMind}
\begin{figure*}[ht]
    \centering
    \includegraphics[width=\linewidth]{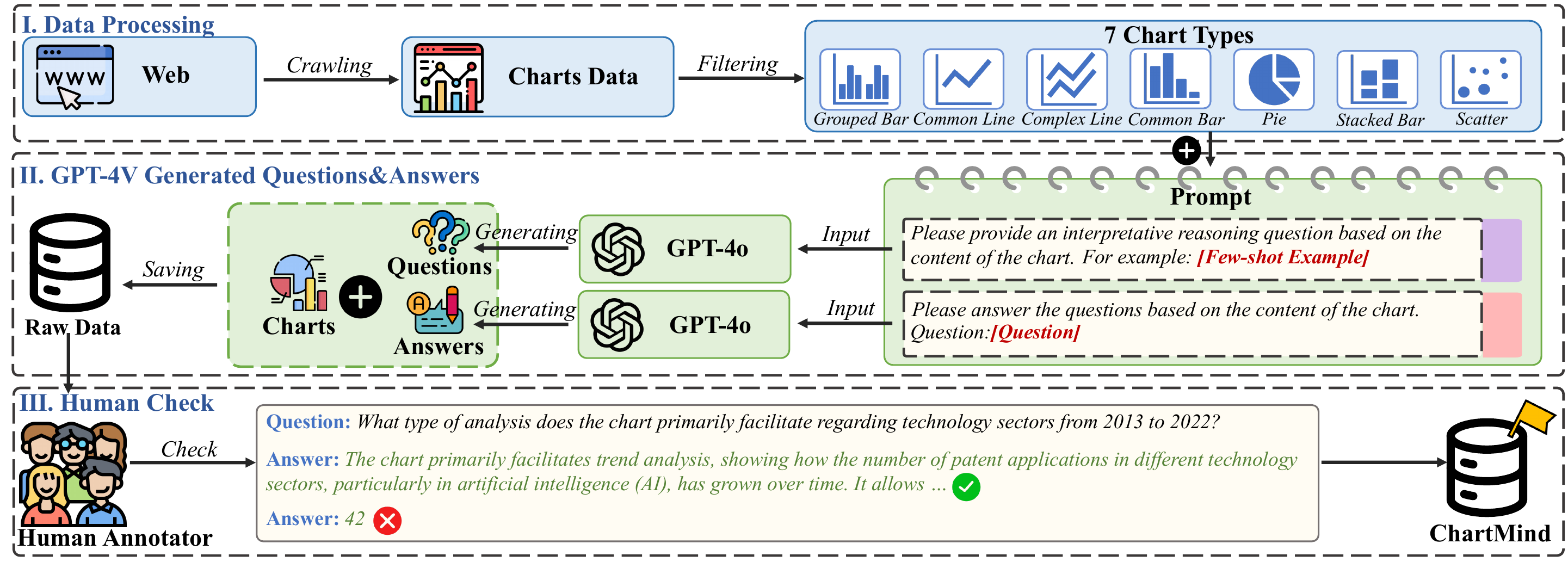} 
    \caption{Data Construction Pipeline for the ChartMind.}
    \label{fig:data_curation}
    \vskip -0.1in
\end{figure*}
Figure~\ref{fig:data_curation} presents an overview of our three-stage data construction pipeline, including chart collection, GPT-based generation, and human validation. Each stage is described below.

\subsection{Stage I: Chart Collection and Processing}
To build a diverse and realistic chart QA benchmark, we collect over 1,200 charts from open-source platforms, including GitHub repositories, public datasets, and Overleaf-based academic projects. All content complies with permissive licenses (e.g., CC BY 4.0, MIT). Charts span multiple formats—pie, bar (common, grouped, stacked), line (common, complex), and scatter plots—covering domains such as economics, education, and technology.

We remove charts that are blurry, lack proper axis or legend labels, or contain unreadable text. This filtering step ensures that remaining charts support meaningful reasoning and are visually accessible to models. These cleaned charts serve as the input to the next stage.

\subsection{Stage II: Prompt-based QA Generation}
Given a chart, we generate diverse QA pairs for seven tasks (e.g., summarization, classification, suggestion) using GPT-4o~\cite{gpt4v}. For each task type, we design a dedicated prompt template that includes a few-shot example, output format instructions, and style control. Prompts are adapted to the chart type and domain to ensure contextual grounding.
To avoid redundancy, we apply controlled randomness (e.g., varying prompt temperature and phrasing) and use clustering on question embeddings to eliminate duplicates. Figure~\ref{fig:data_curation} (Stage II) illustrates this process.

\subsection{Stage III: Human Validation}
Each generated QA pair is reviewed by at least two annotators with over two years of chart QA research experience. Annotators follow a unified protocol and examine: (1) semantic alignment between question and chart, (2) accuracy and consistency of answers, (3) proper use of terminology and metrics. We revise or discard pairs with hallucinated entities, incorrect reasoning, or weak chart grounding.

\paragraph{Answer Rewriting.} GPT-generated answers are not automatically accepted. Annotators verify references to chart elements (e.g., trends, labels, time ranges) and rewrite unclear or incorrect responses. textcolor{blue}{For example:
\textbf{Question:} What does this chart suggest about AI patent trends between 2013 and 2022? 
\textbf{GPT-4o Answer:} They increased significantly. 
\textbf{Human Answer:} The chart shows a consistent rise in AI patent filings, particularly in machine learning, highlighting growing investment in AI research during this period.}

\paragraph{Final Filtering.} Only QA pairs that pass human validation and align with visual evidence are included in ChartMind. Our process draws on best practices from TableBench~\cite{tablebench} and ArXivQA~\cite{arxivqa}. Annotators help refine task definitions by identifying unclear cases.

\subsection{Data Summary and Task Complexity}
\begin{figure}[t]
    \centering
    \includegraphics[width=\linewidth]{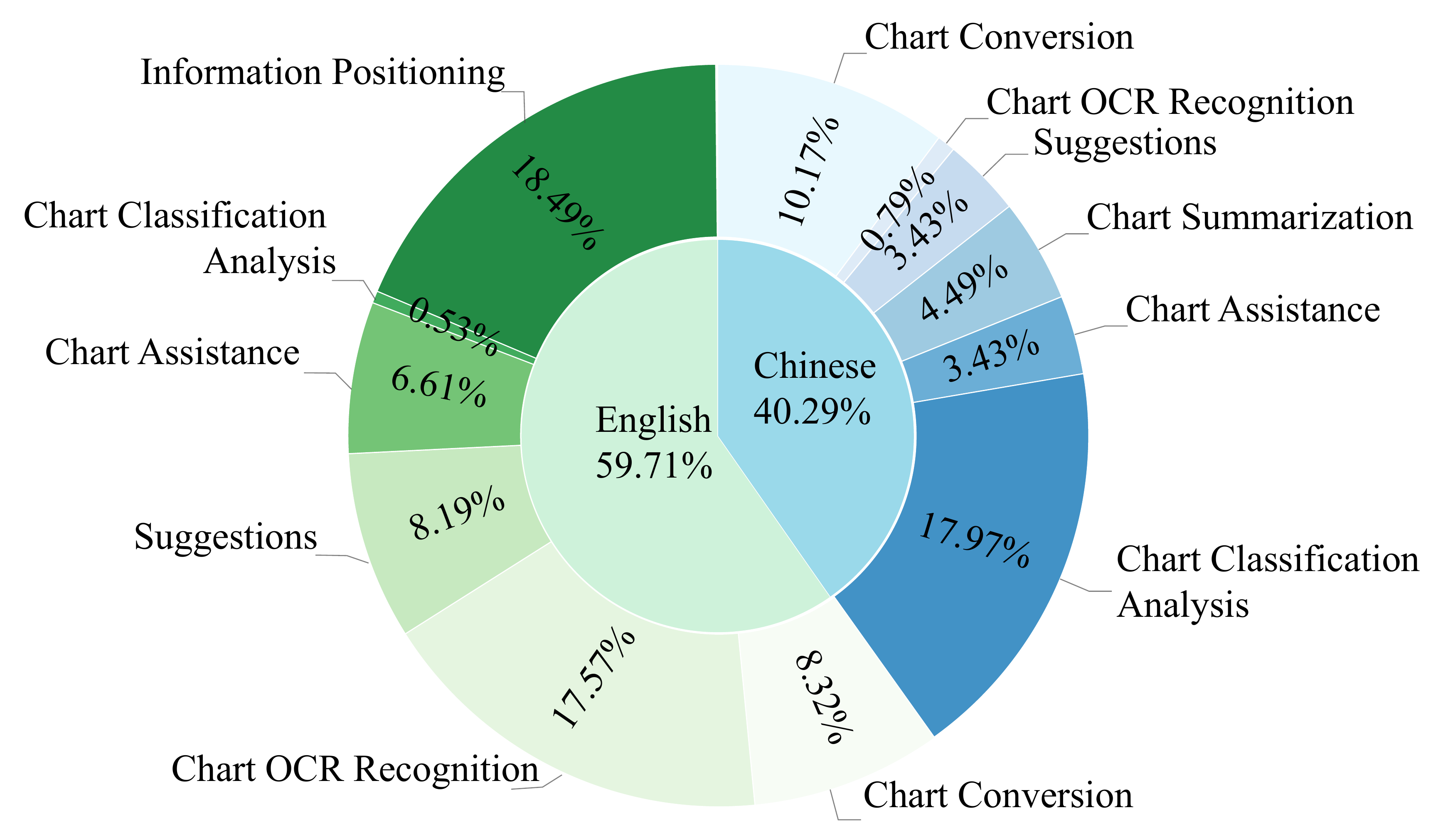}
    \caption{Language and task distribution in ChartMind.}
    \label{fig:task_distribution1}
    \vskip -0.12in
\end{figure}
\begin{figure}[t]
    \centering
    \includegraphics[width=0.82\linewidth]{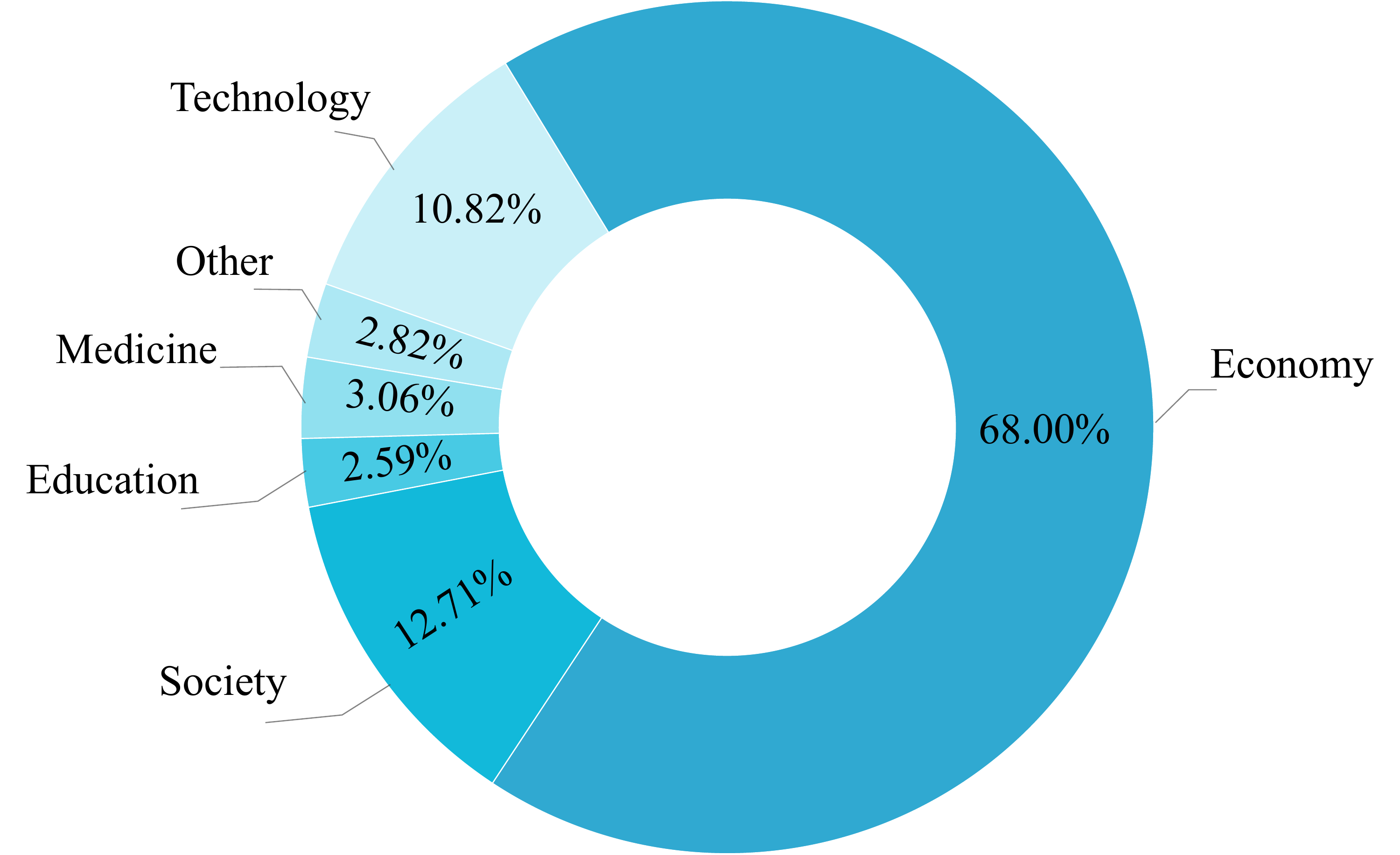}
    \caption{Topic distribution in ChartMind.}
    \label{fig:task_distribution2}
    \vskip -0.12in
\end{figure}

\paragraph{Language and Topic Diversity.} As shown in Figure~\ref{fig:task_distribution1}, ChartMind includes 59.71\% English and 40.29\% Chinese questions, enabling bilingual evaluation across all seven task types. While Chinese is not a low-resource language, high-quality chart reasoning data in Chinese remains rare. ChartMind provides a first step toward multilingual benchmarking, and we plan to expand to more languages in future releases.

Figure~\ref{fig:task_distribution2} illustrates the topic breakdown, where economic charts dominate with 68.00\%, followed by education and technology.

\paragraph{Task Coverage and Reasoning Demands.}
Table~\ref{tab:key_statistics} summarizes the distribution and complexity of QA samples across the seven task categories.

\begin{table}[t]
    \centering
    \resizebox{\linewidth}{!}{
    \begin{tabular}{lcrr}
        \toprule
        \textbf{Task} & \textbf{Samples} & \textbf{Query Length} & \textbf{Answer Length} \\
         &  & \textbf{(Min / Max)} & \textbf{(Min / Max)} \\
        \midrule
        Chart Conversion & 140 & 11 / 477 & 5 / 55 \\
        Chart OCR Recognition & 139 & 13 / 351 & 8 / 59 \\
        Suggestions & 88 & 17 / 492 & 13 / 53 \\
        Chart Classification Analysis & 37 & 360 / 503 & 72 / 79 \\
        Chart Summarization & 34 & 76 / 335 & 12 / 113 \\
        Chart Assistance & 76 & 9 / 276 & 12 / 41 \\
        Information Positioning & 140 & 11 / 208 & 11 / 35 \\
        \midrule
        \textbf{Total} & \textbf{757} & \textbf{9 / 503} & \textbf{5 / 113} \\
        \bottomrule
    \end{tabular}}
    \caption{Task Type Statistics in ChartMind.}
    \label{tab:key_statistics}
    \vspace{-5mm}
\end{table}

The seven task types differ in language structure, visual grounding, and reasoning depth. Summarization and Classification require long, structured responses, while Positioning and OCR involve precise short-form grounding. This diversity supports balanced evaluation of reasoning and generation.

\section{ChartLLM}

\subsection{Problem Definition}
CQA is a task that involves providing an answer \( A \) to a natural language question \( Q \), based on the information contained in a chart \( C \). The answer \( A \) may take various forms, depending on the type of question. Specifically, \( A \) could be a numerical value, a categorical label, an entity set, or an open-domain sentence. These different answer types require distinct reasoning capabilities, ranging from retrieval-based reasoning (e.g., extracting numerical values) to analytical reasoning (e.g., identifying patterns and trends in the chart). Formally, the answer \( A \) is represented as a collection of values or entities \( \{a_1, a_2, \dots, a_k\} \), where \( k \in \mathbb{N}^{+} \).

\subsection{Reasoning Methods}
Instruction-following~\cite{wei2021finetuned} and In-context learning~\cite{dong2024survey} refer to strategies that optimize input for LLMs to generate practical outputs based on task-specific instructions and context. These methods enable models to leverage the provided task instructions to guide reasoning and output generation. To fully assess the reasoning capabilities of LLMs for CQA, we propose three distinct reasoning methods that aim to evaluate the model’s reasoning performance.

\paragraph{Instruction-following-based methods}
Such methods~\cite{wei2021finetuned} leverage task-specific instructions to guide LLMs in reasoning tasks. The model utilizes a prompt to interpret chart data and generate answers. The prompt \( P \) provides additional contextual guidance for the natural language question \( Q \), specifying how the model should reason over the chart data. The reasoning process can be expressed as:
\begin{equation}
M(C, Q, P) \rightarrow A
\end{equation}
where \( M \) represents the model, \( C \) is the chart, \( Q \) is the natural language question, \( P \) is the instruction prompt, and \( A \) is the answer. This approach can be applied in both fine-tuning and zero-shot settings, allowing the model to adapt to tasks based on the provided instructions.

\paragraph{OCR-enhanced methods}
OCR-enhanced methods~\cite{liu2023deplot} augment reasoning by incorporating textual content extracted from charts using OCR tools. These tools provide the model with additional information embedded in the chart, which may not be directly accessible through its visual content. The reasoning process is formulated as:
\begin{equation}
M(C, Q, O(C)) \rightarrow A
\end{equation}
where \( O(C) \) denotes the OCR-extracted content from the chart \( C \). OCR tools offer essential support in understanding chart-based queries by enhancing the model's input with relevant textual data.

\paragraph{COT-based methods}  
COT-based methods~\cite{wei2022chain} break down the reasoning process into intermediate steps to improve both the accuracy and interpretability of the model’s responses. This approach decomposes the reasoning into a sequence of logical steps, which enhances the model’s ability to solve complex tasks. The process is represented as:
\begin{equation}
M(C, Q) \rightarrow \{r_1, r_2, \dots, r_k\} \rightarrow A
\end{equation}
where \( r_1, r_2, \dots, r_k \) represent intermediate reasoning steps, and \( A \) is the final answer. CoT is particularly useful for tasks requiring step-by-step reasoning, such as analyzing trends, identifying patterns, or extracting structured insights from complex chart data.

\begin{table*}[!ht]
  \centering
  \resizebox{\linewidth}{!}{
    \begin{tabular}{lccccccccccr}
    \toprule
    \multirow{2}[4]{*}{ Models} & \multirow{2}[4]{*}{Size} & \multicolumn{3}{c}{ChartMind} & \multicolumn{3}{c}{ChartQA } & \multicolumn{3}{c}{Chart-to-Text} & OpenCQA  \\
\cmidrule{3-12}          &       & ACC & Avg.CIDEr & Avg.GPT-4o Score & Aug. ACC  & Hum. ACC & Avg. ACC & Pew. BLEU & Statista. BLEU & Avg. BLEU & Avg. BLEU \\
    \midrule
    \rowcolor[rgb]{ .922,  .945,  .871} \multicolumn{12}{c}{Instruction-Following-Based~\cite{wei2021finetuned}} \\
    \midrule
    TinyChart\textdagger~\cite{tinychart} & 3B    & 5.36  & 18.45  & 16.81  & 93.60  & 72.16  & 82.88  & 10.84  & 27.04  & 18.94  & 19.62  \\
    ChartInstruct\textdagger~\cite{chartinstruct} & 7B    & 9.82  & 24.55  & 15.05  & 82.40  & 40.64  & 61.52  & 12.81  & 39.39  & 26.10  & 14.78  \\
    ChartLlama\textdagger~\cite{chartllama} & 7B    & 20.54  & 21.34  & 12.72  & 90.36  & 48.96  & 69.66  & 14.23  & 40.71  & 27.47  & 4.70  \\
    Sphinx-v2~\cite{sphinx} & 7B    & 9.82  & 25.95  & 13.69  & 60.96  & 43.92  & 52.44  & 3.43  & 4.94  & 4.19  & 3.10  \\
    LLaVA1.5~\cite{llava1.5} & 7B    & 34.82  & 39.50  & 15.58  & 20.12  & 25.20  & 22.66  & 15.70  & 11.07  & 13.39  & 15.17  \\
    ViP-LLaVA~\cite{vipllava} & 7B    & 20.54  & 37.01  & 15.56  & 17.60  & 26.16  & 21.88  & 1.36  & 2.59  & 1.98  & 15.04  \\
    LLaVA-NEXT~\cite{llavanext} & 7B    & 20.54  & 47.37  & 31.09  & 74.26  & 46.30  & 60.28  & 13.85  & 6.63  & 10.24  & 8.07  \\
    IXC-2.5~\cite{ixc2.5} & 7B    & 47.30  & 40.10  & 43.31  & 92.40  & 74.32  & 83.36  & 17.69  & 11.86  & 14.78  & 9.39  \\
    Qwen2-VL~\cite{qwenvl} & 7B    & 57.14  & 37.32  & 47.89  & 94.10  & 72.00  & 83.05  & 11.07  & 22.98  & 17.03  & 8.26  \\
    mPLUG-Owl2~\cite{mplugowl2} & 8B    & 25.00  & 36.17  & 14.22  & 24.13  & 27.34  & 25.74  & 12.83  & 5.97  & 9.40  & 5.34  \\
    MiniCPM-v2~\cite{minicpmv2} & 8B    & 22.32  & 28.48  & 10.63  & 91.12  & 69.02  & 80.07  & 22.17  & 11.01  & 16.59  & 20.05  \\
    CogVLM~\cite{cogvlm} & 17B   & 23.21  & 40.20  & 29.35  & 23.95  & 39.53  & 31.74  & 16.38  & 11.84  & 14.11  & 1.75  \\
    GLM-4V-plus~\cite{chatglm4v} & - & 59.83  & 38.36  & 21.52  & 16.80  & 12.80  & 14.80  & 5.69  & 5.71  & 5.70  & 7.41  \\
    GPT-4o~\cite{gpt4v} & - & \textbf{61.89}  & 47.25  & 68.81  & 95.34  & 76.06  & 85.70  & 17.75  & 8.70  & 13.23  & 13.92  \\
    \midrule
    \rowcolor[rgb]{ .922,  .945,  .871} \multicolumn{12}{c}{OCR-Enhanced~\cite{liu2023deplot}} \\
    \midrule
    TinyChart\textdagger~\cite{tinychart} & 3B & \makebox[5.4em][s]{6.71 (\textcolor{blue}{+1.35})} & \makebox[5.4em][s]{13.91 (\textcolor{red}{-4.54})} & \makebox[5.4em][s]{17.91 (\textcolor{blue}{+1.10})} & \makebox[5.4em][s]{94.86 (\textcolor{blue}{+1.26})} & \makebox[5.4em][s]{73.95 (\textcolor{blue}{+1.79})} & \makebox[5.4em][s]{84.41 (\textcolor{blue}{+1.53})} & \makebox[5.4em][s]{13.85 (\textcolor{blue}{+3.01})} & \makebox[5.4em][s]{28.27 (\textcolor{blue}{+1.23})} & \makebox[5.4em][s]{21.06 (\textcolor{blue}{+2.12})} & \makebox[5.4em][s]{20.15 (\textcolor{blue}{+0.53})} \\
    ChartInstruct\textdagger~\cite{chartinstruct} & 7B & \makebox[5.4em][s]{{10.01} (\textcolor{blue}{+0.19})} & \makebox[5.4em][s]{32.80 (\textcolor{blue}{+8.25})} & \makebox[5.4em][s]{23.42 (\textcolor{blue}{+8.37})} & \makebox[5.4em][s]{83.74 (\textcolor{blue}{+1.34})} & \makebox[5.4em][s]{42.17 (\textcolor{blue}{+1.53})} & \makebox[5.4em][s]{62.96 (\textcolor{blue}{+1.44})} & \makebox[5.4em][s]{14.95 (\textcolor{blue}{+2.14})} & \makebox[5.4em][s]{40.83 (\textcolor{blue}{+1.44})} & \makebox[5.4em][s]{27.89 (\textcolor{blue}{+1.79})} & \makebox[5.4em][s]{16.01 (\textcolor{blue}{+1.23})} \\
    ChartLlama\textdagger~\cite{chartllama} & 7B & \makebox[5.4em][s]{{22.03} (\textcolor{blue}{+1.49})} & \makebox[5.4em][s]{21.07 (\textcolor{red}{-0.27})} & \makebox[5.4em][s]{26.70 (\textcolor{blue}{+13.97})} & \makebox[5.4em][s]{90.85 (\textcolor{blue}{+0.49})} & \makebox[5.4em][s]{49.26 (\textcolor{blue}{+0.30})} & \makebox[5.4em][s]{70.06 (\textcolor{blue}{+0.40})} & \makebox[5.4em][s]{16.02 (\textcolor{blue}{+1.79})} & \makebox[5.4em][s]{39.97 (\textcolor{red}{-0.74})} & \makebox[5.4em][s]{28.00 (\textcolor{blue}{+0.53})} & \makebox[5.4em][s]{5.89 (\textcolor{blue}{+1.19})} \\
    Sphinx-v2~\cite{sphinx} & 7B & \makebox[5.4em][s]{11.54 (\textcolor{blue}{+1.72})} & \makebox[5.4em][s]{24.14 (\textcolor{red}{-1.81})} & \makebox[5.4em][s]{17.21 (\textcolor{blue}{+3.52})} & \makebox[5.4em][s]{64.08 (\textcolor{blue}{+3.12})} & \makebox[5.4em][s]{45.49 (\textcolor{blue}{+1.57})} & \makebox[5.4em][s]{54.79 (\textcolor{blue}{+2.35})} & \makebox[5.4em][s]{8.81 (\textcolor{blue}{+5.38})} & \makebox[5.4em][s]{2.39 (\textcolor{red}{-2.55})} & \makebox[5.4em][s]{5.60 (\textcolor{blue}{+1.41})} & \makebox[5.4em][s]{3.16 (\textcolor{blue}{+0.06})} \\
    LLaVA1.5~\cite{llava1.5} & 7B & \makebox[5.4em][s]{36.15 (\textcolor{blue}{+1.33})} & \makebox[5.4em][s]{33.49 (\textcolor{red}{-6.01})} & \makebox[5.4em][s]{21.03 (\textcolor{blue}{+5.45})} & \makebox[5.4em][s]{19.73 (\textcolor{red}{-0.39})} & \makebox[5.4em][s]{25.95 (\textcolor{blue}{+0.75})} & \makebox[5.4em][s]{22.84 (\textcolor{blue}{+0.18})} & \makebox[5.4em][s]{15.94 (\textcolor{blue}{+0.24})} & \makebox[5.4em][s]{12.67 (\textcolor{blue}{+1.60})} & \makebox[5.4em][s]{14.30 (\textcolor{blue}{+0.91})} & \makebox[5.4em][s]{16.31 (\textcolor{blue}{+1.14})} \\
    ViP-LLaVA~\cite{vipllava} & 7B & \makebox[5.4em][s]{25.38 (\textcolor{blue}{+4.84})} & \makebox[5.4em][s]{36.77 (\textcolor{red}{-0.24})} & \makebox[5.4em][s]{26.45 (\textcolor{blue}{+10.89})} & \makebox[5.4em][s]{27.12 (\textcolor{blue}{+9.52})} & \makebox[5.4em][s]{24.94 (\textcolor{red}{-1.22})} & \makebox[5.4em][s]{26.03 (\textcolor{blue}{+4.15})} & \makebox[5.4em][s]{14.13 (\textcolor{blue}{+12.77})} & \makebox[5.4em][s]{14.37 (\textcolor{blue}{+11.78})} & \makebox[5.4em][s]{14.25 (\textcolor{blue}{+12.27})} & \makebox[5.4em][s]{18.08 (\textcolor{blue}{+3.04})} \\
    LLaVA-NEXT~\cite{llavanext} & 7B & \makebox[5.4em][s]{41.15 (\textcolor{blue}{+20.61})} & \makebox[5.4em][s]{47.83 (\textcolor{blue}{+0.46})} & \makebox[5.4em][s]{31.51 (\textcolor{blue}{+0.42})} & \makebox[5.4em][s]{70.47 (\textcolor{red}{-3.79})} & \makebox[5.4em][s]{52.68 (\textcolor{blue}{+6.38})} & \makebox[5.4em][s]{61.58 (\textcolor{blue}{+1.30})} & \makebox[5.4em][s]{15.16 (\textcolor{blue}{+1.31})} & \makebox[5.4em][s]{8.82 (\textcolor{blue}{+2.19})} & \makebox[5.4em][s]{11.99 (\textcolor{blue}{+1.75})} & \makebox[5.4em][s]{8.25 (\textcolor{blue}{+0.18})} \\
    IXC-2.5~\cite{ixc2.5} & 7B & \makebox[5.4em][s]{42.31 (\textcolor{red}{-4.99})} & \makebox[5.4em][s]{40.35 (\textcolor{blue}{+0.24})} & \makebox[5.4em][s]{45.38 (\textcolor{blue}{+2.06})} & \makebox[5.4em][s]{94.23 (\textcolor{blue}{+1.83})} & \makebox[5.4em][s]{73.40 (\textcolor{red}{-0.92})} & \makebox[5.4em][s]{83.82 (\textcolor{blue}{+0.46})} & \makebox[5.4em][s]{17.03 (\textcolor{red}{-0.66})} & \makebox[5.4em][s]{12.34 (\textcolor{blue}{+0.48})} & \makebox[5.4em][s]{14.68 (\textcolor{red}{-0.10})} & \makebox[5.4em][s]{14.53 (\textcolor{blue}{+5.14})} \\
    Qwen2-VL~\cite{qwenvl} & 7B & \makebox[5.4em][s]{42.31 (\textcolor{red}{-14.83})} & \makebox[5.4em][s]{36.04 (\textcolor{red}{-1.27})} & \makebox[5.4em][s]{49.28 (\textcolor{blue}{+1.39})} & \makebox[5.4em][s]{94.23 (\textcolor{blue}{+0.13})} & \makebox[5.4em][s]{75.96 (\textcolor{blue}{+3.96})} & \makebox[5.4em][s]{85.10 (\textcolor{blue}{+2.05})} & \makebox[5.4em][s]{11.08 (\textcolor{blue}{+0.01})} & \makebox[5.4em][s]{23.21 (\textcolor{blue}{+0.23})} & \makebox[5.4em][s]{17.15 (\textcolor{blue}{+0.12})} & \makebox[5.4em][s]{11.75 (\textcolor{blue}{+3.49})} \\
    mPLUG-Owl2~\cite{mplugowl2} & 8B & \makebox[5.4em][s]{27.62 (\textcolor{blue}{+2.62})} & \makebox[5.4em][s]{30.60 (\textcolor{red}{-5.57})} & \makebox[5.4em][s]{24.67 (\textcolor{blue}{+10.44})} & \makebox[5.4em][s]{35.58 (\textcolor{blue}{+11.45})} & \makebox[5.4em][s]{37.18 (\textcolor{blue}{+9.84})} & \makebox[5.4em][s]{36.38 (\textcolor{blue}{+10.65})} & \makebox[5.4em][s]{11.82 (\textcolor{red}{-1.01})} & \makebox[5.4em][s]{7.30 (\textcolor{blue}{+1.33})} & \makebox[5.4em][s]{9.56 (\textcolor{blue}{+0.16})} & \makebox[5.4em][s]{4.45 (\textcolor{red}{-0.89})} \\
    MiniCPM-v2~\cite{minicpmv2} & 8B & \makebox[5.4em][s]{23.04 (\textcolor{blue}{+0.72}} & \makebox[5.4em][s]{19.73 (\textcolor{red}{-8.75})} & \makebox[5.4em][s]{18.10 (\textcolor{blue}{+7.47})} & \makebox[5.4em][s]{\makebox[5.4em][s]{92.36 (\textcolor{blue}{+1.24})}} & \makebox[5.4em][s]{73.21 (\textcolor{blue}{+4.19})} & \makebox[5.4em][s]{82.79 (\textcolor{blue}{+2.72})} & \makebox[5.4em][s]{20.93 (\textcolor{red}{-1.24})} & \makebox[5.4em][s]{5.75 (\textcolor{red}{-5.26})} & \makebox[5.4em][s]{13.34 (\textcolor{red}{-3.25})} & \makebox[5.4em][s]{20.60 (\textcolor{blue}{+0.55})} \\
    CogVLM~\cite{cogvlm} & 17B & \makebox[5.4em][s]{25.54 (\textcolor{blue}{+2.33})} & \makebox[5.4em][s]{39.00 (\textcolor{red}{-1.20})} & \makebox[5.4em][s]{36.80 (\textcolor{blue}{+7.45})} & \makebox[5.4em][s]{29.81 (\textcolor{blue}{+5.86})} & \makebox[5.4em][s]{48.72 (\textcolor{blue}{+9.19})} & \makebox[5.4em][s]{39.27 (\textcolor{blue}{+7.53})} & \makebox[5.4em][s]{20.85 (\textcolor{blue}{+4.47})} & \makebox[5.4em][s]{13.88 (\textcolor{blue}{+2.04})} & \makebox[5.4em][s]{17.37 (\textcolor{blue}{+3.26})} & \makebox[5.4em][s]{1.79 (\textcolor{blue}{+0.04})} \\
    GLM-4V-plus~\cite{chatglm4v} & - & \makebox[5.4em][s]{44.64 (\textcolor{red}{-15.19})} & \makebox[5.4em][s]{44.83 (\textcolor{blue}{+6.47})} & \makebox[5.4em][s]{35.79 (\textcolor{blue}{+14.27})} & \makebox[5.4em][s]{17.95 (\textcolor{blue}{+1.15})} & \makebox[5.4em][s]{16.87 (\textcolor{blue}{+4.07})} & \makebox[5.4em][s]{17.41 (\textcolor{blue}{+2.61})} & \makebox[5.4em][s]{7.91 (\textcolor{blue}{+2.22})} & \makebox[5.4em][s]{7.63 (\textcolor{blue}{+1.92})} & \makebox[5.4em][s]{7.77 (\textcolor{blue}{+2.07})} & \makebox[5.4em][s]{8.72 (\textcolor{blue}{+1.31})} \\
    GPT-4o~\cite{gpt4v} & - & \makebox[5.4em][s]{49.31 (\textcolor{red}{-12.58})} & \makebox[5.4em][s]{46.48 (\textcolor{red}{-0.76})} & \makebox[5.4em][s]{\underline{71.79} (\textcolor{blue}{+2.98})} & \makebox[5.4em][s]{\underline{96.20} (\textcolor{blue}{+0.86})} & \makebox[5.4em][s]{\underline{78.04} (\textcolor{blue}{+1.98})} & \makebox[5.4em][s]{\underline{87.12} (\textcolor{blue}{+1.42})} & \makebox[5.4em][s]{20.13 (\textcolor{blue}{+2.38})} & \makebox[5.4em][s]{9.86 (\textcolor{blue}{+1.16})} & \makebox[5.4em][s]{15.00 (\textcolor{blue}{+1.77})} & \makebox[5.4em][s]{14.85 (\textcolor{blue}{+0.93})} \\
    \midrule
    \rowcolor[rgb]{ .922,  .945,  .871} \multicolumn{12}{c}{COT-Based~\cite{wei2022chain}} \\
    \midrule
    TinyChart\textdagger~\cite{tinychart} & 3B & \makebox[5.4em][s]{6.01 (\textcolor{blue}{+0.65})} & \makebox[5.4em][s]{13.58 (\textcolor{red}{-4.87})} & \makebox[5.4em][s]{19.30 (\textcolor{blue}{+2.49})} & \makebox[5.4em][s]{94.84 (\textcolor{blue}{+1.24})} & \makebox[5.4em][s]{74.46 (\textcolor{blue}{+2.30})} & \makebox[5.4em][s]{84.65 (\textcolor{blue}{+1.77})} & \makebox[5.4em][s]{12.31 (\textcolor{blue}{+1.47})} & \makebox[5.4em][s]{28.53 (\textcolor{blue}{+1.49})} & \makebox[5.4em][s]{20.42 (\textcolor{blue}{+1.48})} & \makebox[5.4em][s]{20.74 (\textcolor{blue}{+1.12})} \\
    ChartInstruct\textdagger~\cite{chartinstruct} & 7B & \makebox[5.4em][s]{9.96 (\textcolor{blue}{+0.14})} & \makebox[5.4em][s]{31.95 (\textcolor{blue}{+7.40})} & \makebox[5.4em][s]{22.44 (\textcolor{blue}{+7.39})} & \makebox[5.4em][s]{83.35 (\textcolor{blue}{+0.95})} & \makebox[5.4em][s]{42.74 (\textcolor{blue}{+2.10})} & \makebox[5.4em][s]{63.05 (\textcolor{blue}{+1.53})} & \makebox[5.4em][s]{14.34 (\textcolor{blue}{+1.53})} & \makebox[5.4em][s]{41.32 (\textcolor{blue}{+1.93})} & \makebox[5.4em][s]{27.83 (\textcolor{blue}{+1.73})} & \makebox[5.4em][s]{15.25 (\textcolor{blue}{+0.47})} \\
    ChartLlama\textdagger~\cite{chartllama} & 7B & \makebox[5.4em][s]{21.44 (\textcolor{blue}{+0.90})} & \makebox[5.4em][s]{18.99 (\textcolor{red}{-2.36})} & \makebox[5.4em][s]{21.77 (\textcolor{blue}{+9.04})} & \makebox[5.4em][s]{91.63 (\textcolor{blue}{+1.27})} & \makebox[5.4em][s]{50.04 (\textcolor{blue}{+1.08})} & \makebox[5.4em][s]{70.84 (\textcolor{blue}{+1.18})} & \makebox[5.4em][s]{15.76 (\textcolor{blue}{+1.53})} & \makebox[5.4em][s]{\underline{41.42} (\textcolor{blue}{+0.71})} & \makebox[5.4em][s]{\underline{28.59} (\textcolor{blue}{+1.12})} & \makebox[5.4em][s]{6.32 (\textcolor{blue}{+1.62})} \\
    Sphinx-v2~\cite{sphinx} & 7B & \makebox[5.4em][s]{9.91 (\textcolor{blue}{+0.09})} & \makebox[5.4em][s]{25.03 (\textcolor{red}{-0.92})} & \makebox[5.4em][s]{16.26 (\textcolor{blue}{+2.57})} & \makebox[5.4em][s]{61.86 (\textcolor{blue}{+0.90})} & \makebox[5.4em][s]{46.79 (\textcolor{blue}{+2.87})} & \makebox[5.4em][s]{54.33 (\textcolor{blue}{+1.89})} & \makebox[5.4em][s]{3.53 (\textcolor{blue}{+0.10})} & \makebox[5.4em][s]{5.09 (\textcolor{blue}{+0.15})} & \makebox[5.4em][s]{4.31 (\textcolor{blue}{+0.12})} & \makebox[5.4em][s]{3.13 (\textcolor{blue}{+0.03})} \\
    LLaVA1.5~\cite{llava1.5} & 7B & \makebox[5.4em][s]{35.77 (\textcolor{blue}{+0.95})} & \makebox[5.4em][s]{35.61 (\textcolor{red}{-3.89})} & \makebox[5.4em][s]{19.68 (\textcolor{blue}{+4.10})} & \makebox[5.4em][s]{16.90 (\textcolor{red}{-3.22})} & \makebox[5.4em][s]{28.57 (\textcolor{blue}{+3.37})} & \makebox[5.4em][s]{22.74 (\textcolor{blue}{+0.08})} & \makebox[5.4em][s]{15.20 (\textcolor{red}{-0.50})} & \makebox[5.4em][s]{11.66 (\textcolor{blue}{+0.59})} & \makebox[5.4em][s]{13.43 (\textcolor{blue}{+0.04})} & \makebox[5.4em][s]{15.93 (\textcolor{blue}{+0.76})} \\
    ViP-LLaVA~\cite{vipllava} & 7B & \makebox[5.4em][s]{23.31 (\textcolor{blue}{+2.77})} & \makebox[5.4em][s]{36.13 (\textcolor{red}{-0.88})} & \makebox[5.4em][s]{22.24 (\textcolor{blue}{+6.68})} & \makebox[5.4em][s]{22.12 (\textcolor{blue}{+4.52})} & \makebox[5.4em][s]{28.21 (\textcolor{blue}{+2.05})} & \makebox[5.4em][s]{25.17 (\textcolor{blue}{+3.29})} & \makebox[5.4em][s]{15.48 (\textcolor{blue}{+14.12})} & \makebox[5.4em][s]{12.20 (\textcolor{blue}{+9.61})} & \makebox[5.4em][s]{13.84 (\textcolor{blue}{+11.86})} & \makebox[5.4em][s]{15.67 (\textcolor{blue}{+0.63})} \\
    LLaVA-NEXT~\cite{llavanext} & 7B & \makebox[5.4em][s]{40.23 (\textcolor{blue}{+19.69})} & \makebox[5.4em][s]{47.44 (\textcolor{blue}{+0.07})} & \makebox[5.4em][s]{27.34 (\textcolor{red}{-3.75})} & \makebox[5.4em][s]{68.49 (\textcolor{red}{-5.77})} & \makebox[5.4em][s]{52.13 (\textcolor{blue}{+5.83})} & \makebox[5.4em][s]{60.31 (\textcolor{blue}{+0.03})} & \makebox[5.4em][s]{14.81 (\textcolor{blue}{+0.96})} & \makebox[5.4em][s]{6.29 (\textcolor{red}{-0.34})} & \makebox[5.4em][s]{10.55 (\textcolor{blue}{+0.31})} & \makebox[5.4em][s]{8.09 (\textcolor{blue}{+0.02})} \\
    IXC-2.5~\cite{ixc2.5} & 7B & \makebox[5.4em][s]{41.15 (\textcolor{red}{-6.15})} & \makebox[5.4em][s]{41.23 (\textcolor{blue}{+1.13})} & \makebox[5.4em][s]{46.73 (\textcolor{blue}{+3.42})} & \makebox[5.4em][s]{93.91 (\textcolor{blue}{+1.51})} & \makebox[5.4em][s]{72.82 (\textcolor{red}{-1.50})} & \makebox[5.4em][s]{83.37 (\textcolor{blue}{+0.01})} & \makebox[5.4em][s]{17.36 (\textcolor{blue}{+0.23})} & \makebox[5.4em][s]{11.92 (\textcolor{blue}{+0.06})} & \makebox[5.4em][s]{14.64 (\textcolor{red}{-0.14})} & \makebox[5.4em][s]{14.39 (\textcolor{blue}{+5.00})} \\
    Qwen2-VL~\cite{qwenvl} & 7B & \makebox[5.4em][s]{40.69 (\textcolor{red}{-16.45})} & \makebox[5.4em][s]{44.72 (\textcolor{blue}{+7.41})} & \makebox[5.4em][s]{55.12 (\textcolor{blue}{+7.24})} & \makebox[5.4em][s]{94.87 (\textcolor{blue}{+0.77})} & \makebox[5.4em][s]{77.88 (\textcolor{blue}{+5.88})} & \makebox[5.4em][s]{86.38 (\textcolor{blue}{+3.33})} & \makebox[5.4em][s]{16.70 (\textcolor{blue}{+5.63})} & \makebox[5.4em][s]{23.91 (\textcolor{blue}{+0.93})} & \makebox[5.4em][s]{20.30 (\textcolor{blue}{+3.27})} & \makebox[5.4em][s]{10.32 (\textcolor{blue}{+2.06})} \\
    mPLUG-Owl2~\cite{mplugowl2} & 8B & \makebox[5.4em][s]{25.89 (\textcolor{blue}{+0.89})} & \makebox[5.4em][s]{35.10 (\textcolor{red}{-1.08})} & \makebox[5.4em][s]{21.27 (\textcolor{blue}{+7.04})} & \makebox[5.4em][s]{27.56 (\textcolor{blue}{+3.43})} & \makebox[5.4em][s]{31.09 (\textcolor{blue}{+3.75})} & \makebox[5.4em][s]{29.33 (\textcolor{blue}{+3.59})} & \makebox[5.4em][s]{14.00 (\textcolor{blue}{+1.17})} & \makebox[5.4em][s]{7.84 (\textcolor{blue}{+1.87})} & \makebox[5.4em][s]{10.92 (\textcolor{blue}{+1.52})} & \makebox[5.4em][s]{7.88 (\textcolor{blue}{+2.54})} \\
    MiniCPM-v2~\cite{minicpmv2} & 8B & \makebox[5.4em][s]{22.78 (\textcolor{blue}{+0.46})} & \makebox[5.4em][s]{28.81 (\textcolor{blue}{+0.33})} & \makebox[5.4em][s]{18.18 (\textcolor{blue}{+7.54})} &  \makebox[5.4em][s]{92.37 (\textcolor{blue}{+1.25})} & \makebox[5.4em][s]{71.47 (\textcolor{blue}{+2.45})} & \makebox[5.4em][s]{81.92 (\textcolor{blue}{+1.85})} & \makebox[5.4em][s]{\underline{26.56} (\textcolor{blue}{+4.39})} & \makebox[5.4em][s]{12.53 (\textcolor{blue}{+1.52})} & \makebox[5.4em][s]{19.54 (\textcolor{blue}{+2.95})} & \makebox[5.4em][s]{20.30 (\textcolor{blue}{+0.25})} \\
    CogVLM~\cite{cogvlm} & 17B & \makebox[5.4em][s]{24.01 (\textcolor{blue}{+0.80})} & \makebox[5.4em][s]{40.04 (\textcolor{red}{-0.16})} & \makebox[5.4em][s]{37.14 (\textcolor{blue}{+7.79})} & \makebox[5.4em][s]{27.31 (\textcolor{blue}{+3.36})} & \makebox[5.4em][s]{44.93 (\textcolor{blue}{+5.40})} & \makebox[5.4em][s]{36.12 (\textcolor{blue}{+4.38})} & \makebox[5.4em][s]{17.94 (\textcolor{blue}{+1.56})} & \makebox[5.4em][s]{12.57 (\textcolor{blue}{+0.73})} & \makebox[5.4em][s]{15.26 (\textcolor{blue}{+1.15})} & \makebox[5.4em][s]{3.41 (\textcolor{blue}{+1.66})} \\
    GLM-4V-plus~\cite{chatglm4v} & - & \makebox[5.4em][s]{41.00 (\textcolor{red}{-18.83})} & \makebox[5.4em][s]{39.55 (\textcolor{blue}{+1.19})} & \makebox[5.4em][s]{21.68 (\textcolor{blue}{+0.16})} & \makebox[5.4em][s]{18.63 (\textcolor{blue}{+1.83})} & \makebox[5.4em][s]{15.96 (\textcolor{blue}{+3.16})} & \makebox[5.4em][s]{17.30 (\textcolor{blue}{+2.50})} & \makebox[5.4em][s]{6.86 (\textcolor{blue}{+1.17})} & \makebox[5.4em][s]{7.72 (\textcolor{blue}{+2.01})} & \makebox[5.4em][s]{7.29 (\textcolor{blue}{+1.59})} & \makebox[5.4em][s]{8.83 (\textcolor{blue}{+1.42})} \\
    GPT-4o~\cite{gpt4v} & - & \makebox[5.4em][s]{46.15 (\textcolor{red}{-15.74})} & \makebox[5.4em][s]{\underline{48.19} (\textcolor{blue}{+0.95})} & \makebox[5.4em][s]{69.00 (\textcolor{blue}{+0.19})} & \makebox[5.4em][s]{95.39 (\textcolor{blue}{+0.05})} & \makebox[5.4em][s]{77.23 (\textcolor{blue}{+1.17})} & \makebox[5.4em][s]{86.31 (\textcolor{blue}{+0.61})} & \makebox[5.4em][s]{19.20 (\textcolor{blue}{+1.45})} & \makebox[5.4em][s]{9.31 (\textcolor{blue}{+0.61})} & \makebox[5.4em][s]{14.26 (\textcolor{blue}{+1.03})} & \makebox[5.4em][s]{15.42 (\textcolor{blue}{+1.50})} \\
    \midrule
    \rowcolor[rgb]{ .922,  .945,  .871} \multicolumn{12}{c}{ChartLLM-Based} \\
    \midrule
    TinyChart\textdagger~\cite{tinychart} & 3B & \makebox[5.4em][s]{7.69 (\textcolor{blue}{+2.33})} & \makebox[5.4em][s]{20.07 (\textcolor{blue}{+1.62})} & \makebox[5.4em][s]{23.21 (\textcolor{blue}{+6.40})} & \makebox[5.4em][s]{95.04 (\textcolor{blue}{+1.44})} & \makebox[5.4em][s]{74.41 (\textcolor{blue}{+2.25})} & \makebox[5.4em][s]{84.73 (\textcolor{blue}{+1.85})} & \makebox[5.4em][s]{14.68 (\textcolor{blue}{+3.84})} & \makebox[5.4em][s]{34.22 (\textcolor{blue}{+7.18})} & \makebox[5.4em][s]{24.45 (\textcolor{blue}{+5.51})} & \makebox[5.4em][s]{\textbf{21.84 } (\textcolor{blue}{+2.22})} \\
    ChartInstruct\textdagger~\cite{chartinstruct} & 7B & \makebox[5.4em][s]{11.54 (\textcolor{blue}{+1.72})} & \makebox[5.4em][s]{34.79 (\textcolor{blue}{+10.24})} & \makebox[5.4em][s]{26.43 (\textcolor{blue}{+11.39})} & \makebox[5.4em][s]{85.93 (\textcolor{blue}{+3.53})} & \makebox[5.4em][s]{43.52 (\textcolor{blue}{+2.88})} & \makebox[5.4em][s]{64.73 (\textcolor{blue}{+3.20})} & \makebox[5.4em][s]{15.52 (\textcolor{blue}{+2.71})} & \makebox[5.4em][s]{\underline{41.42} (\textcolor{blue}{+2.03})} & \makebox[5.4em][s]{28.47 (\textcolor{blue}{+2.37})} & \makebox[5.4em][s]{18.53 (\textcolor{blue}{+3.75})} \\
    ChartLlama\textdagger~\cite{chartllama} & 7B & \makebox[5.4em][s]{22.67 (\textcolor{blue}{+2.13})} & \makebox[5.4em][s]{22.54 (\textcolor{blue}{+1.19})} & \makebox[5.4em][s]{27.58 (\textcolor{blue}{+14.85})} & \makebox[5.4em][s]{91.42 (\textcolor{blue}{+1.06})} & \makebox[5.4em][s]{51.72 (\textcolor{blue}{+2.76})} & \makebox[5.4em][s]{71.57 (\textcolor{blue}{+1.91})} & \makebox[5.4em][s]{17.94 (\textcolor{blue}{+3.71})} & \makebox[5.4em][s]{\textbf{40.47 } (\textcolor{red}{-0.24})} & \makebox[5.4em][s]{\textbf{29.21 } (\textcolor{blue}{+1.74})} & \makebox[5.4em][s]{7.40 (\textcolor{blue}{+2.70})} \\
    Sphinx-v2~\cite{sphinx} & 7B & \makebox[5.4em][s]{13.85 (\textcolor{blue}{+4.03})} & \makebox[5.4em][s]{30.11 (\textcolor{blue}{+4.16})} & \makebox[5.4em][s]{23.68 (\textcolor{blue}{+9.99})} & \makebox[5.4em][s]{62.80 (\textcolor{blue}{+1.84})} & \makebox[5.4em][s]{48.00 (\textcolor{blue}{+4.08})} & \makebox[5.4em][s]{55.40 (\textcolor{blue}{+2.96})} & \makebox[5.4em][s]{7.90 (\textcolor{blue}{+4.47})} & \makebox[5.4em][s]{7.35 (\textcolor{blue}{+2.41})} & \makebox[5.4em][s]{7.63 (\textcolor{blue}{+3.44})} & \makebox[5.4em][s]{6.88 (\textcolor{blue}{+3.78})} \\
    LLaVA1.5~\cite{llava1.5} & 7B & \makebox[5.4em][s]{36.92 (\textcolor{blue}{+2.10})} & \makebox[5.4em][s]{38.39 (\textcolor{red}{-1.11})} & \makebox[5.4em][s]{26.95 (\textcolor{blue}{+11.37})} & \makebox[5.4em][s]{25.44 (\textcolor{blue}{+5.32})} & \makebox[5.4em][s]{31.68 (\textcolor{blue}{+6.48})} & \makebox[5.4em][s]{28.56 (\textcolor{blue}{+5.90})} & \makebox[5.4em][s]{18.21 (\textcolor{blue}{+2.51})} & \makebox[5.4em][s]{17.83 (\textcolor{blue}{+6.76})} & \makebox[5.4em][s]{18.02 (\textcolor{blue}{+4.63})} & \makebox[5.4em][s]{17.40 (\textcolor{blue}{+2.23})} \\
    ViP-LLaVA~\cite{vipllava} & 7B & \makebox[5.4em][s]{26.23 (\textcolor{blue}{+5.69})} & \makebox[5.4em][s]{41.98 (\textcolor{blue}{+4.97})} & \makebox[5.4em][s]{28.79 (\textcolor{blue}{+13.23})} & \makebox[5.4em][s]{23.96 (\textcolor{blue}{+6.36})} & \makebox[5.4em][s]{29.04 (\textcolor{blue}{+2.88})} & \makebox[5.4em][s]{26.50 (\textcolor{blue}{+4.62})} & \makebox[5.4em][s]{14.31 (\textcolor{blue}{+12.95})} & \makebox[5.4em][s]{14.38 (\textcolor{blue}{+11.79})} & \makebox[5.4em][s]{14.35 (\textcolor{blue}{+12.37})} & \makebox[5.4em][s]{18.72 (\textcolor{blue}{+3.68})} \\
    LLaVA-NEXT~\cite{llavanext} & 7B & \makebox[5.4em][s]{42.31 (\textcolor{blue}{+21.77})} & \makebox[5.4em][s]{49.40 (\textcolor{blue}{+2.03})} & \makebox[5.4em][s]{34.40 (\textcolor{blue}{+3.32})} & \makebox[5.4em][s]{75.82 (\textcolor{blue}{+1.56})} & \makebox[5.4em][s]{47.68 (\textcolor{blue}{+1.38})} & \makebox[5.4em][s]{61.75 (\textcolor{blue}{+1.47})} & \makebox[5.4em][s]{15.26 (\textcolor{blue}{+1.41})} & \makebox[5.4em][s]{8.93 (\textcolor{blue}{+2.30})} & \makebox[5.4em][s]{12.10 (\textcolor{blue}{+1.86})} & \makebox[5.4em][s]{9.02 (\textcolor{blue}{+0.95})} \\
    IXC-2.5~\cite{ixc2.5} & 7B & \makebox[5.4em][s]{47.31 (\textcolor{blue}{+0.01})} & \makebox[5.4em][s]{43.38 (\textcolor{blue}{+3.28})} & \makebox[5.4em][s]{51.88 (\textcolor{blue}{+8.56})} & \makebox[5.4em][s]{94.88 (\textcolor{blue}{+2.48})} & \makebox[5.4em][s]{76.24 (\textcolor{blue}{+1.92})} & \makebox[5.4em][s]{85.56 (\textcolor{blue}{+2.20})} & \makebox[5.4em][s]{19.82 (\textcolor{blue}{+2.13})} & \makebox[5.4em][s]{14.70 (\textcolor{blue}{+2.84})} & \makebox[5.4em][s]{17.26 (\textcolor{blue}{+2.48})} & \makebox[5.4em][s]{16.83 (\textcolor{blue}{+7.44})} \\
    Qwen2-VL~\cite{qwenvl} & 7B & \makebox[5.4em][s]{57.66 (\textcolor{blue}{+0.52})} & \makebox[5.4em][s]{45.54 (\textcolor{blue}{+8.22})} & \makebox[5.4em][s]{56.10 (\textcolor{blue}{+8.21})} & \makebox[5.4em][s]{94.40 (\textcolor{blue}{+0.30})} & \makebox[5.4em][s]{77.44 (\textcolor{blue}{+5.44})} & \makebox[5.4em][s]{85.92 (\textcolor{blue}{+2.87})} & \makebox[5.4em][s]{20.96 (\textcolor{blue}{+9.89})} & \makebox[5.4em][s]{24.45 (\textcolor{blue}{+1.47})} & \makebox[5.4em][s]{22.71 (\textcolor{blue}{+5.68})} & \makebox[5.4em][s]{18.53 (\textcolor{blue}{+10.27})} \\
    mPLUG-Owl2~\cite{mplugowl2} & 8B & \makebox[5.4em][s]{29.38 (\textcolor{blue}{+4.38})} & \makebox[5.4em][s]{40.46 (\textcolor{blue}{+4.29})} & \makebox[5.4em][s]{29.15 (\textcolor{blue}{+14.93})} & \makebox[5.4em][s]{38.76 (\textcolor{blue}{+14.63})} & \makebox[5.4em][s]{40.34 (\textcolor{blue}{+13.00})} & \makebox[5.4em][s]{39.55 (\textcolor{blue}{+13.82})} & \makebox[5.4em][s]{13.01 (\textcolor{blue}{+0.18})} & \makebox[5.4em][s]{8.91 (\textcolor{blue}{+2.94})} & \makebox[5.4em][s]{10.96 (\textcolor{blue}{+1.56})} & \makebox[5.4em][s]{6.26 (\textcolor{blue}{+0.92})} \\
    MiniCPM-v2~\cite{minicpmv2} & 8B & \makebox[5.4em][s]{24.21 (\textcolor{blue}{+1.89})} & \makebox[5.4em][s]{38.65 (\textcolor{blue}{+10.17})} & \makebox[5.4em][s]{23.73 (\textcolor{blue}{+13.09})} & \makebox[5.4em][s]{93.84 (\textcolor{blue}{+2.72})} & \makebox[5.4em][s]{71.86 (\textcolor{blue}{+2.84})} & \makebox[5.4em][s]{82.85 (\textcolor{blue}{+2.78})} & \makebox[5.4em][s]{\textbf{27.68 } (\textcolor{blue}{+5.51})} & \makebox[5.4em][s]{24.55 (\textcolor{blue}{+13.54})} & \makebox[5.4em][s]{26.12 (\textcolor{blue}{+9.53})} & \makebox[5.4em][s]{\underline{20.88} (\textcolor{blue}{+0.83})} \\
    CogVLM~\cite{cogvlm} & 17B & \makebox[5.4em][s]{26.38 (\textcolor{blue}{+3.17})} & \makebox[5.4em][s]{41.05 (\textcolor{blue}{+0.85})} & \makebox[5.4em][s]{41.85 (\textcolor{blue}{+12.50})} & \makebox[5.4em][s]{33.41 (\textcolor{blue}{+9.46})} & \makebox[5.4em][s]{51.73 (\textcolor{blue}{+12.20})} & \makebox[5.4em][s]{42.57 (\textcolor{blue}{+10.83})} & \makebox[5.4em][s]{21.46 (\textcolor{blue}{+5.08})} & \makebox[5.4em][s]{14.74 (\textcolor{blue}{+2.90})} & \makebox[5.4em][s]{18.10 (\textcolor{blue}{+3.99})} & \makebox[5.4em][s]{2.48 (\textcolor{blue}{+0.73})} \\
    GLM-4V-plus~\cite{chatglm4v} & - & \makebox[5.4em][s]{\underline{60.18} (\textcolor{blue}{+0.35})} & \makebox[5.4em][s]{47.00 (\textcolor{blue}{+8.64})} & \makebox[5.4em][s]{37.19 (\textcolor{blue}{+15.67})} & \makebox[5.4em][s]{19.74 (\textcolor{blue}{+2.94})} & \makebox[5.4em][s]{18.04 (\textcolor{blue}{+5.24})} & \makebox[5.4em][s]{19.66 (\textcolor{blue}{+4.86})} & \makebox[5.4em][s]{9.75 (\textcolor{blue}{+4.06})} & \makebox[5.4em][s]{8.97 (\textcolor{blue}{+3.26})} & \makebox[5.4em][s]{9.36 (\textcolor{blue}{+3.66})} & \makebox[5.4em][s]{9.74 (\textcolor{blue}{+2.34})} \\
    GPT-4o~\cite{gpt4v} & - & \makebox[5.4em][s]{\textbf{61.89 } (\textcolor{blue}{+0.00})} & \makebox[5.4em][s]{\textbf{50.42 } (\textcolor{blue}{+3.17})} & \makebox[5.4em][s]{\textbf{73.89 } (\textcolor{blue}{+5.08})} & \makebox[5.4em][s]{{\textbf{98.63 }} (\textcolor{blue}{+3.29})} & \makebox[5.4em][s]{\textbf{79.49 } (\textcolor{blue}{+3.43})} & \makebox[5.4em][s]{\textbf{89.06 } (\textcolor{blue}{+3.36})} & \makebox[5.4em][s]{23.65 (\textcolor{blue}{+5.90})} & \makebox[5.4em][s]{11.07 (\textcolor{blue}{+2.37})} & \makebox[5.4em][s]{17.36 (\textcolor{blue}{+4.14})} & \makebox[5.4em][s]{16.04 (\textcolor{blue}{+2.12})} \\
    \bottomrule
    \end{tabular}
    }
    \raggedright
    \caption{Performance of multimodal models on ChartMind and three structured-output CQA datasets. The best results are highlighted in \textbf{bold}, and the second-best results are \underline{underlined}. \textdagger Specialized CQA models.}
    \vskip -0.15in
  \label{tab:results1}
\end{table*}

\subsection{ChartLLM: Context Extraction for CQA}

The ChartLLM is designed to enhance CQA by extracting and structuring relevant contextual information from a chart. Given a chart \( C \), the context \( C_{\text{context}} = \{ T, L, X, Y \} \), where \( T \) is the title, \( L \) is the legend, \( X \) is the X-axis label, and \( Y \) is the Y-axis label, is generated to represent the essential elements of the chart. This approach minimizes irrelevant data and focuses solely on the components required for accurate reasoning in CQA tasks. To extract \( C_{\text{context}} \), predefined prompts, such as "Extract key information from the chart, including title, legend, and X and Y-axis information," guide the model in identifying the necessary elements of the chart. 
This ensures the extracted context is concise, relevant, and foundational for reasoning. Unlike step-by-step CoT reasoning, ChartLLM focuses on structured context modeling, reducing the model’s perceptual burden by presenting semantically key components upfront.

The reasoning objective for ChartLLM is to predict the answer \( A \) that maximizes the conditional probability given the question \( Q \) and the extracted context \( C_{\text{context}} \). This can be expressed as:
\begin{equation}
\resizebox{\linewidth}{!}{$
A = \operatorname{argmax}_{a \in \mathcal{A}} 
\sum_{i=1}^{n} \mathbb{E}_{C_{\text{context}}, Q} 
\left[ \log P(a_i \mid C_{\text{context}}, Q; \Theta) \right]
$}
\label{eq:chartllm_objective}
\end{equation}
Here, \( A \) is the predicted answer, \( \mathcal{A} \) represents the candidate answer space, \( C_{\text{context}} \) is the extracted context from the chart \( C \), \( Q \) is the natural language question, \( a_i \) is the \( i \)-th candidate answer, and \( \Theta \) denotes the model parameters.


\section{Experiments}

\subsection{Experimental Setup}
We evaluate four paradigms for CQA tasks, including instruction-following, COT-based reasoning, OCR-enhanced methods, and our proposed ChartLLM framework. These methods are tested on 14 MLLMs from three categories: specialized CQA models, general-purpose open-source models, and general-purpose closed-source models. The evaluation spans four datasets, including our proposed ChartMind and three structured-output CQA datasets—ChartQA~\cite{chartqa}, Chart-to-Text~\cite{charttotext}, and OpenCQA~\cite{opencqa}—which primarily rely on predefined answer formats and automated scoring metrics. In contrast, ChartMind introduces diverse chart formats and open-domain textual outputs, enabling a more comprehensive assessment of real-world CQA scenarios. Further implementation details, model descriptions, and benchmark specifications are provided in Appendix~\ref{appendix:exp_setup}.

\begin{table}[t]
\centering
\resizebox{\linewidth}{!}{
\begin{tabular}{lccc}
\toprule
\textbf{Models} & \textbf{Size} & \textbf{Avg. GPT-4o Score} & \textbf{Avg. Human Score} \\
\midrule
ChartInstruct~\cite{chartinstruct} & 7B & 26.43 & 22.52 \\
ChartLlama~\cite{chartllama} & 7B & 27.58 & 23.11 \\
TinyChart~\cite{tinychart} & 3B & 23.21 & 21.97 \\
mPLUG-Owl2~\cite{mplugowl2} & 8B & 29.15 & 29.31 \\
Sphinx-v2~\cite{sphinx} & 7B & 23.68 & 22.31 \\
CogVLM~\cite{cogvlm} & 17B & 41.85 & 34.96 \\
LLaVA1.5~\cite{llava1.5} & 7B & 26.95 & 22.93 \\
MiniCPM-v2~\cite{minicpmv2} & 8B & 23.73 & 24.01 \\
ViP-LLaVA~\cite{vipllava} & 7B & 28.79 & 30.75 \\
LLaVA-NEXT~\cite{llavanext} & 7B & 34.40 & 32.31 \\
IXC-2.5~\cite{ixc2.5} & 7B & 51.88 & 36.61 \\
Qwen2-VL~\cite{qwenvl} & 7B & 56.10 & 40.39 \\
GLM-4V-plus~\cite{chatglm4v} & - & 37.19 & 39.35 \\
GPT-4o~\cite{gpt4v} & - & 73.89 & 50.73 \\
\midrule
\textbf{PCC~\cite{cohen2009pearson}} & - & \multicolumn{2}{c}{\textbf{93.09}} \\
\bottomrule
\end{tabular}}
\caption{Correlation of GPT4o and Human Eval.}
\vskip -0.2in
\label{tab:consistency_metrics}
\end{table}

\subsection{Main Results}  
To evaluate the effectiveness and robustness of ChartLLM-based methods over OCR-enhanced~\cite{liu2023deplot} and COT-based~\cite{wei2022chain} approaches in open-ended and structured-output reasoning, \autoref{tab:results1} compares their performance across various benchmarks. Both OCR-enhanced and COT-based methods yield significant improvements (blue text), but their effectiveness varies by task. OCR-enhanced methods often degrade performance (red text), particularly in open-ended reasoning, where redundancy and noise from textual extraction disrupt holistic reasoning. For instance, GPT-4o’s~\cite{gpt4v} ACC in open-ended tasks drops by -12.58 with OCR-enhanced methods, reflecting their sensitivity to flexible reasoning. COT-based methods enhance structured-output reasoning but struggle in open-ended tasks, reducing GPT-4o’s ACC by -15.74 due to difficulties in integrating contextual and visual elements. ChartLLM-based methods address these challenges by strategically extracting key contextual information and minimizing redundancy, reducing external noise in reasoning. By focusing on essential chart elements and preserving relevant semantic relationships, they achieve superior performance with consistent adaptability across both reasoning types. Their ability to balance context extraction and noise reduction underscores their robustness in handling complex chart reasoning.

\subsection{Correlation Analysis of Metrics}
To assess the consistency between automated and human evaluation in open-ended CQA, \autoref{tab:consistency_metrics} analyzes the correlation between GPT-4o Score and Human Score across 14 multimodal models. The Pearson Correlation Coefficient (PCC)~\cite{cohen2009pearson} is 93.09, indicating a strong linear relationship. High-performing models like GPT-4o~\cite{gpt4v} and Qwen2-VL~\cite{qwenvl} show strong alignment between GPT-4o and human scores, validating automated evaluation reliability. Notably, models like mPLUG-Owl2~\cite{mplugowl2} and ViP-LLaVA~\cite{vipllava} exhibit slight deviations, where human scores marginally exceed automated ones, possibly reflecting nuanced human judgment in open-ended reasoning. The high PCC confirms GPT-4o Score as a robust proxy for human evaluation, reinforcing its applicability in open-ended CQA.

\subsection{Sensitivity Analysis}
\paragraph{Language-Level Analysis.}
\begin{figure}[t]
    \centering
    \includegraphics[width=\linewidth, height=0.38\textheight]{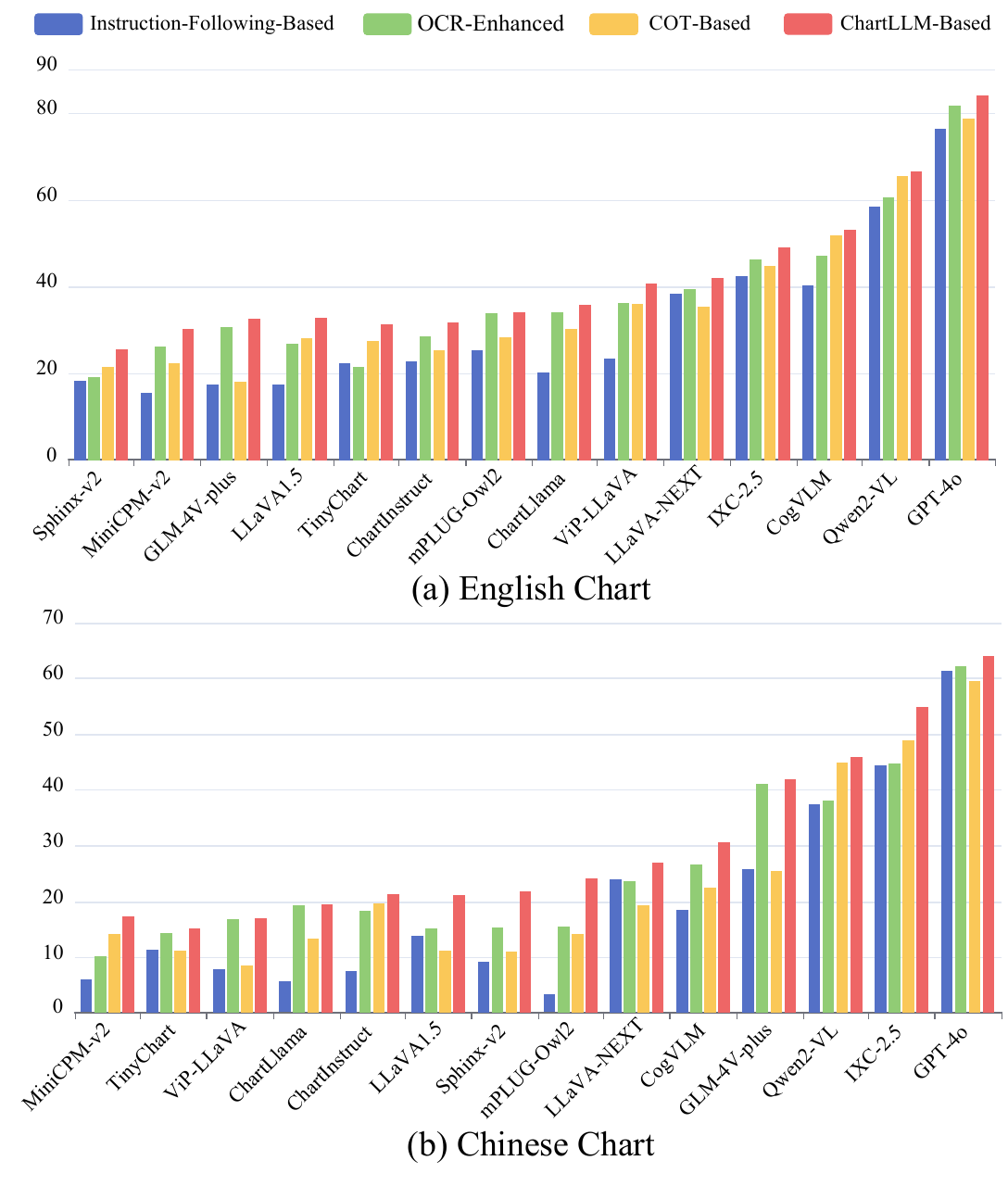}
    \caption{Performance of multimodal models across Chinese and English datasets in ChartMind.}
    \label{fig:gpt4o_evaluation2}
    \vskip -0.15in
\end{figure}
\begin{figure*}[t]
    \centering
    \includegraphics[width=\linewidth]{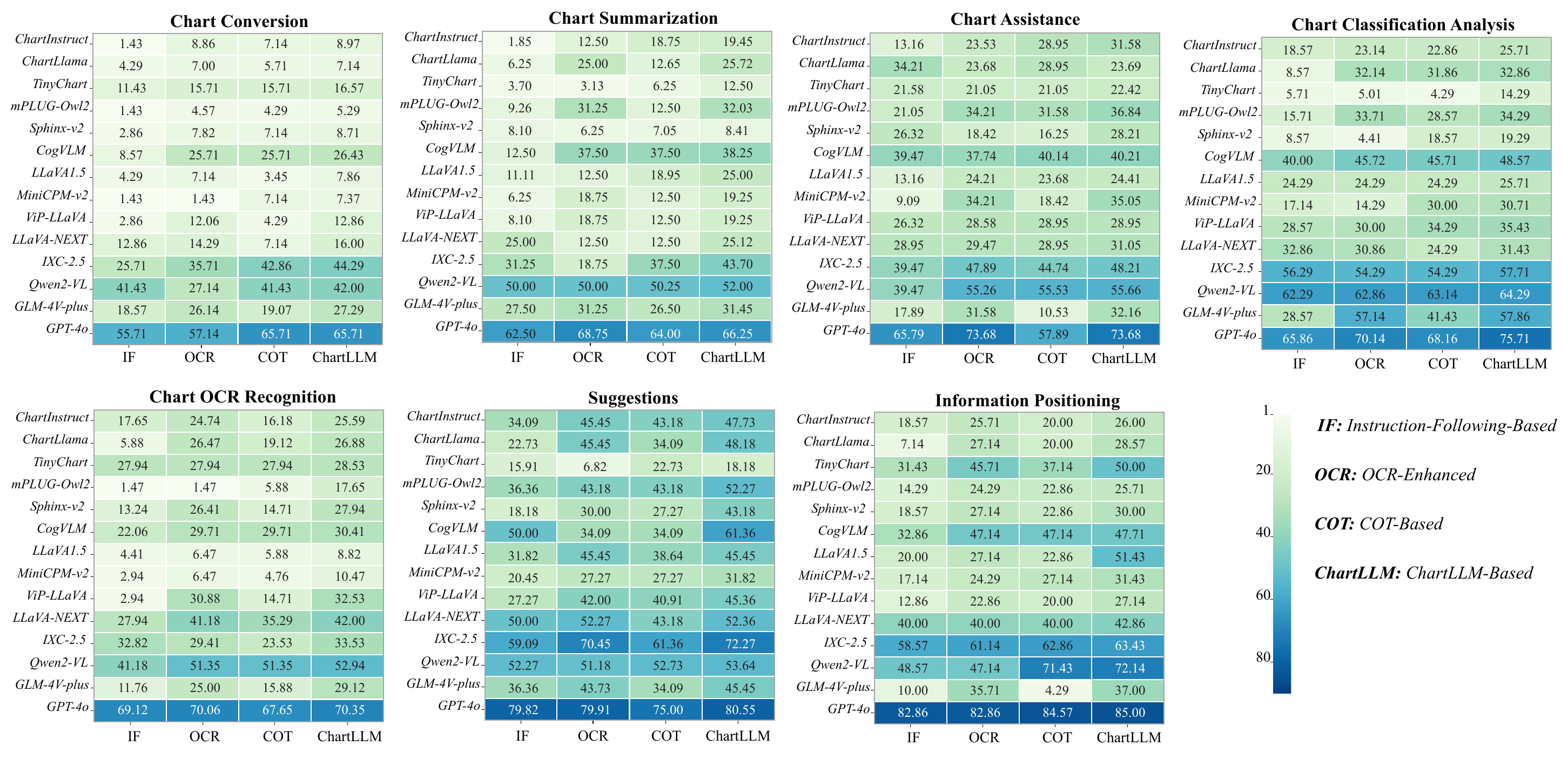}
    \vskip -0.1in
    \caption{Performance of multimodal models on seven tasks in ChartMind.}
    \label{fig:gpt4o_evaluation1}
    \vskip -0.15in
\end{figure*}

To evaluate the sensitivity of different paradigms to multilingual challenges in CQA, we analyze model performance across English and Chinese charts in ChartMind. Figure~\ref{fig:gpt4o_evaluation2} compares results under each method across both languages, grouped by paradigm to highlight method robustness. We observe a consistent performance gap across models: Chinese tasks are generally more difficult, reflecting challenges in tokenization, OCR quality, and implicit reasoning common in Chinese chart labels. Instruction-following models such as GPT-4o~\cite{gpt4v} and LLaVA1.5~\cite{llava1.5} show significant degradation in Chinese due to weaker multilingual grounding.
OCR-enhanced methods help mitigate these gaps by injecting extracted text, especially in Chinese, where axis labels and titles are often more semantically informative. COT-based methods help slightly but introduce more variance, especially in visual tasks where decomposition is less intuitive.
ChartLLM-based methods consistently achieve the best cross-lingual performance. By explicitly structuring chart context before reasoning, ChartLLM reduces noise and enhances semantic alignment, leading to more stable performance in both languages.

\paragraph{Task-Level Analysis.}

To explore how different paradigms handle diverse CQA tasks, we evaluate model performance across seven task types in ChartMind. As shown in \autoref{fig:gpt4o_evaluation1}, these tasks vary in difficulty. \textit{Chart Conversion} and \textit{Chart Summarization} are the most challenging, involving semantic fusion and cross-modal reasoning. In contrast, \textit{Suggestions} and \textit{Information Positioning} focus on localized extraction and are comparatively easier.
Instruction-following methods often struggle with complex tasks, showing unstable outputs due to weak multimodal alignment. OCR-enhanced approaches perform well in text-heavy scenarios like \textit{Chart OCR Recognition}, but degrade on tasks such as \textit{Summarization}, where excess raw text introduces noise and misleads the model.
COT-based methods help in procedural reasoning tasks like \textit{Suggestions}, but fall short in integrative tasks such as \textit{Chart Assistance}, where linear step-by-step thinking cannot capture multimodal dependencies.
ChartLLM-based methods consistently demonstrate robust performance across all task types. By explicitly modeling structural context before reasoning, ChartLLM improves semantic grounding in complex settings while preserving precision in simpler tasks. This balance highlights its adaptability and makes it particularly effective for real-world CQA.

\paragraph{Chart-Type-Level Analysis.}
We examine how different paradigms perform across chart types of varying complexity in ChartMind. Tasks involving \textit{Pie} and \textit{Stacked Bar} charts require high-context reasoning, while \textit{Complex Line} charts mainly involve direct value extraction.
Instruction-following models struggle with layout-heavy formats; OCR-enhanced methods perform well on text-dense charts but falter when visual cues dominate. COT-based methods show moderate stability but lack semantic depth.
ChartLLM consistently outperforms others by explicitly modeling contextual elements, enabling it to generalize across both visually intricate and text-sparse chart types. A full breakdown is provided in~\autoref{appendix:chart_type_analysis}.

\section{Conclusion}
We introduce \textit{ChartMind}, the first benchmark for complex CQA in realistic settings. It addresses key gaps in prior work by supporting multilingual charts, open-ended outputs, and seven distinct task types. 
Across four paradigms and 14 multimodal models, our results show that ChartLLM—a model-agnostic, context-aware framework—consistently outperforms OCR and CoT methods, establishing a strong baseline for future CQA research.
Future work will explore multi-turn dialogues, cross-chart reasoning, and hybrid chart–text queries to support more advanced and realistic use cases.

\section*{Limitations}
ChartMind provides a benchmark for complex CQA evaluation, yet several limitations remain. First, the dataset primarily relies on publicly available charts, potentially introducing biases in data distribution and task complexity. Ensuring broader representativeness requires further dataset expansion and diversification. Second, although ChartMind defines seven reasoning tasks, real-world chart analysis often involves more advanced reasoning, such as multi-turn interactions, cross-chart comparisons, and textual-visual information integration, which remain underexplored. Third, the reliance on automated evaluation methods, such as GPT-4 ratings, introduces challenges in capturing nuanced human judgment in complex reasoning. Addressing these issues requires refining evaluation methodologies and incorporating more human annotations. Future improvements may focus on expanding the dataset, enhancing evaluation metrics, and integrating multi-turn reasoning and cross-chart analysis to better reflect real-world scenarios.

\bibliography{custom}

\appendix
\section{Chart Types and Tasks in ChartMind}
\begin{figure*}[t]
    \centering
    \includegraphics[width=\linewidth]{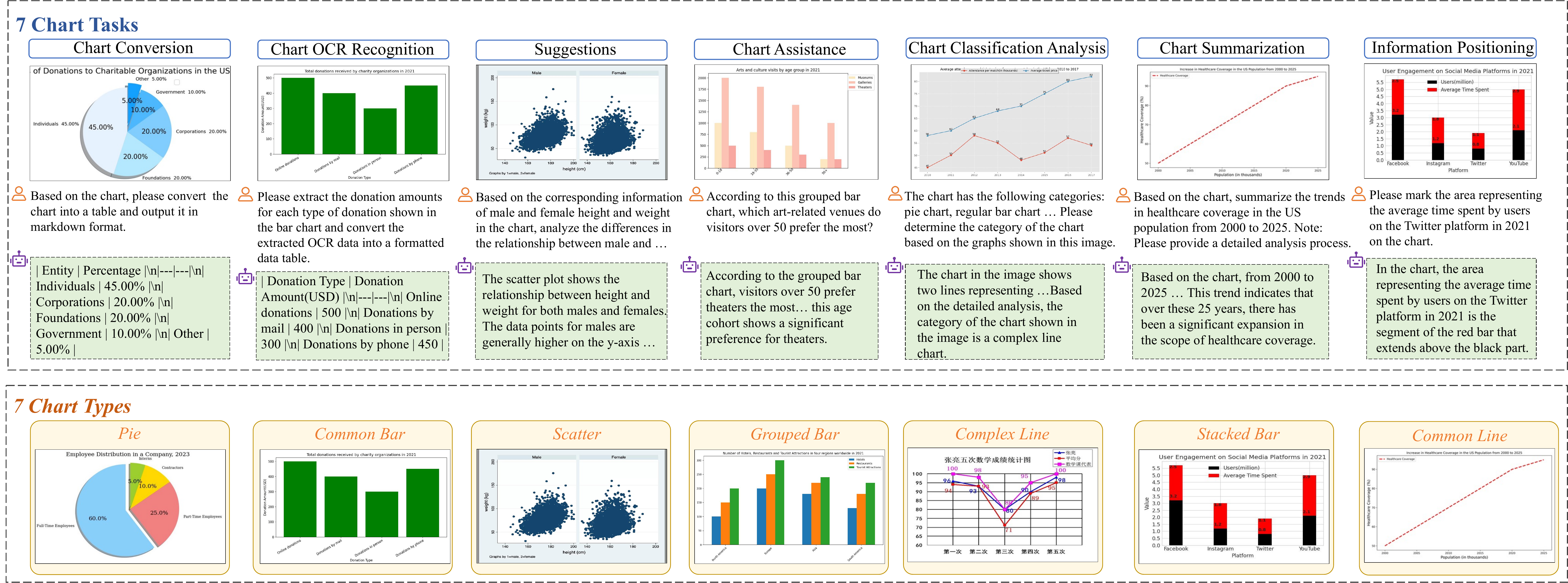}
    \caption{Overview of the seven chart types and seven reasoning tasks included in ChartMind.}
    \label{fig:chart_types_tasks}
\end{figure*}
ChartMind supports a diverse range of chart types and reasoning tasks, ensuring a comprehensive evaluation of complex reasoning in CQA. As shown in \autoref{fig:chart_types_tasks} The dataset includes seven distinct chart types—Pie, Common Bar, Scatter, Grouped Bar, Complex Line, Stacked Bar, and Common Line—capturing varied visual structures and data representations. Additionally, ChartMind defines seven reasoning tasks: Chart Conversion, Chart OCR Recognition, Suggestions, Chart Assistance, Chart Classification, Chart Summarization, and Information Positioning, covering key aspects of multimodal chart understanding. These distributions illustrate ChartMind’s ability to comprehensively assess complex multimodal reasoning, spanning diverse chart types and reasoning paradigms. Compared to prior benchmarks, ChartMind provides a broader evaluation scope, capturing the complexity of real-world CQA tasks.

\section{Experimental Setup Details}
\label{appendix:exp_setup}

\subsection{Implementation Details}
To assess the performance of models on complex CQA tasks in real-world settings, we experiment with four types of paradigms. First, we test MLLMs in the instruction-following setting~\cite{zhou2023instruction}, where we use prompts to evaluate their ability to answer chart-related questions. Second, we apply COT-based methods~\cite{wei2022chain}, which break down reasoning processes into intermediate steps to generate answers. Third, we adopt OCR-enhanced methods inspired by DePlot~\cite{liu2023deplot}, which extract chart content as text and use it as input for multimodal reasoning models. Finally, we propose the ChartLLM method, which enhances reasoning performance by extracting structured contextual information, such as chart titles, legends, and axes, using Qwen2-VL~\cite{qwenvl}, and feeding this information into models for further analysis.

\subsection{Models}
We evaluate 14 MLLMs across three categories: specialized CQA models, general-purpose open-source multimodal models, and general-purpose closed-source multimodal models. The majority of the models have a parameter size of approximately 7B, with a few exceptions, including smaller models such as TinyChart~\cite{tinychart} with 3B parameters and larger models like CogVLM~\cite{cogvlm} with 17B parameters. For specialized CQA models, we include ChartInstruct~\cite{chartinstruct}, ChartLlama~\cite{chartllama}, and TinyChart~\cite{tinychart}. These models are specifically trained on CQA datasets, making them particularly suited for tasks requiring precise understanding of chart-related queries. Among open-source general-purpose multimodal models, we evaluate mPLUG-Owl2~\cite{mplugowl2}, Sphinx-v2~\cite{sphinx}, CogVLM~\cite{cogvlm}, LLaVA1.5~\cite{llava1.5}, MiniCPM-v2~\cite{minicpmv2}, ViP-LLaVA~\cite{vipllava}, LLaVA-NEXT~\cite{llavanext}, IXC-2.5~\cite{ixc2.5}, and Qwen2-VL~\cite{qwenvl}. These models leverage extensive multimodal training datasets, including CQA data, and exhibit strong performance on chart-related tasks. Finally, closed-source general multimodal models, including GPT-4o~\cite{gpt4v} and GLM-4V-plus~\cite{chatglm4v}, are state-of-the-art models with advanced multimodal reasoning capacities, providing strong competition to existing open-source systems.

\begin{figure*}[ht]
    \centering
    \includegraphics[width=\linewidth]{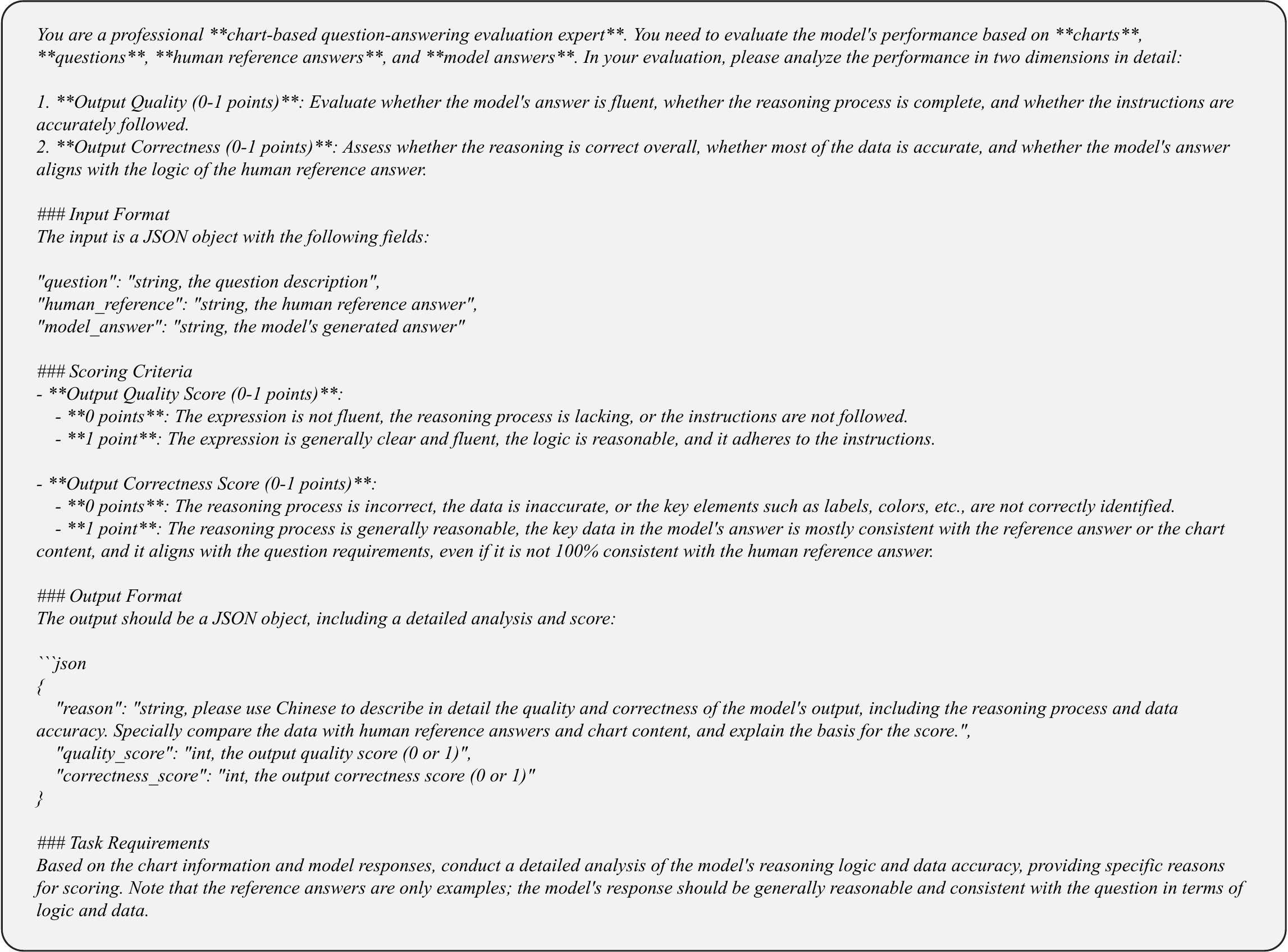}
    \caption{Prompt design for GPT-4o score.}
    \label{fig:gpt4o_evaluation}
\end{figure*}

\subsection{Benchmarks and Metrics}
To comprehensively evaluate multimodal CQA tasks, we adopt three representative structured-output reasoning datasets—ChartQA~\cite{chartqa}, Chart-to-Text~\cite{charttotext}, and OpenCQA~\cite{opencqa}—alongside our proposed benchmark, ChartMind. ChartQA and Chart-to-Text primarily take a chart and a natural language question as input and generate structured textual answers, such as numerical values, categorical labels, or predefined captions, making them well-suited for factual extraction tasks. OpenCQA, despite allowing open-ended queries, constrains responses to structured formats evaluated by automated metrics like BLEU, limiting its ability to assess flexible reasoning. To address these constraints, ChartMind introduces a more comprehensive evaluation by supporting diverse chart types, open-ended textual outputs, and seven complex reasoning tasks, enabling a broader assessment of multimodal reasoning. Models are evaluated using Accuracy and CIDEr for structured assessments, while GPT-4o score and Human score serve as open-ended evaluation metrics, with GPT-4o score as the primary metric, as detailed in ~\autoref{appendixa.1}. The structured-output datasets are evaluated using Accuracy and BLEU score.

\section{GPT-4o Scoring Prompt Design}
\label{appendixa.1}
The GPT-4o score prompt evaluates the performance of models on CQA tasks by assessing two key dimensions: output quality and output correctness. Output quality focuses on the fluency of the model's answer, the completeness of its reasoning process, and its ability to follow instructions accurately. Output correctness measures the overall accuracy of the reasoning, the correctness of the data, and the logical alignment with the human reference answer or chart content. The input to the prompt includes a JSON object containing the question, the human reference answer, and the model-generated answer. The output is also formatted as a JSON object, which includes a detailed explanation of the scoring rationale along with scores for both dimensions. The full design of the scoring prompt is visualized in ~\autoref{fig:gpt4o_evaluation}.

\section{Chart-Type-Level Analysis}
\label{appendix:chart_type_analysis}
\begin{figure*}[ht]
    \centering
    \includegraphics[width=\textwidth]{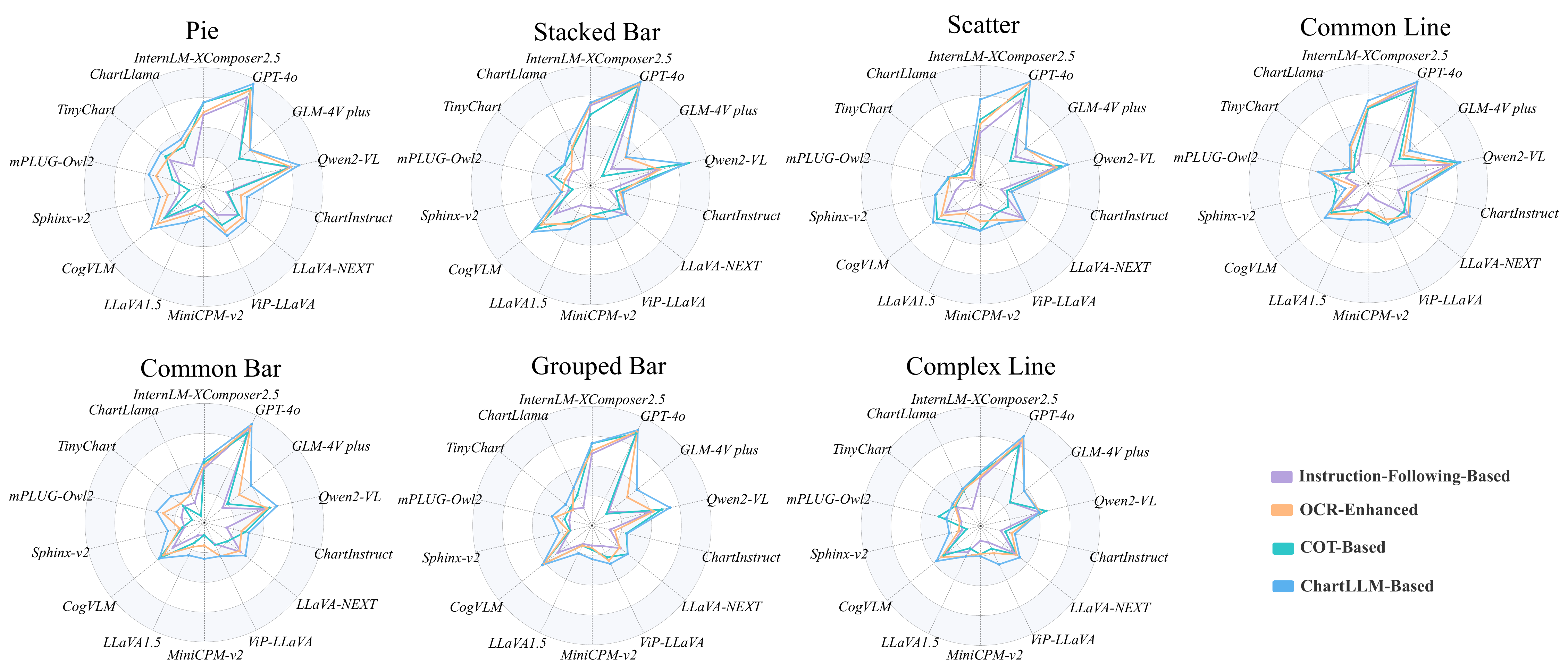} 
    \caption{Performance of multimodal models across chart types, categorized by four paradigms.}
    \label{fig:chart_type_analysis}
\end{figure*}
\begin{figure*}[ht]
    \centering
    \includegraphics[width=\linewidth]{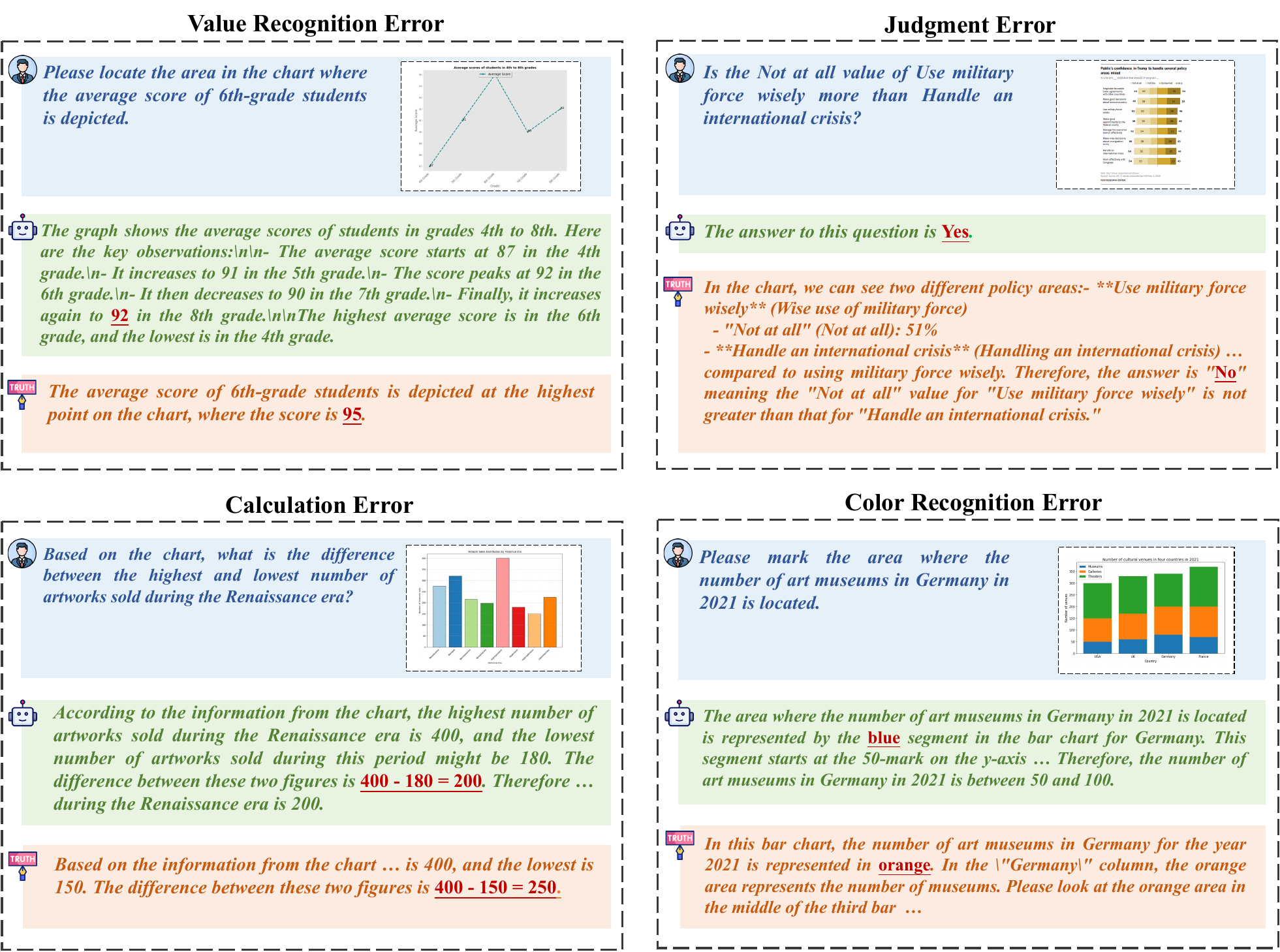}
    \caption{The four major error types in ChartMind.}
    \label{fig:error_analysis}
\end{figure*}
To evaluate the sensitivity of different paradigms to diverse chart types in CQA tasks, we analyze their performance across seven chart types in ChartMind. \autoref{fig:chart_type_analysis} presents a detailed breakdown of model performance. Chart types exhibit varying complexity, with \textit{Pie} and \textit{Stacked Bar} being the most challenging due to their reliance on integrated contextual reasoning, while simpler types like \textit{Complex Line} primarily require straightforward data extraction. Instruction-following methods~\cite{wei2021finetuned}, such as GPT-4o~\cite{gpt4v} and LLAVA1.5~\cite{llava1.5}, show significant performance drops in high-complexity charts, underscoring their limitations in managing holistic reasoning tasks. OCR-enhanced methods~\cite{liu2023deplot} excel in text-heavy charts such as \textit{Grouped Bar}, leveraging their ability to extract textual information, but struggle with tasks like \textit{Scatter} that demand comprehensive visual-semantic integration. COT-based methods~\cite{wei2022chain} demonstrate moderate performance across most chart types, performing relatively well in structured charts like \textit{Common Line}, yet falling short in tasks requiring high-contextual reasoning. ChartLLM-based methods achieve the highest overall performance, excelling in high-difficulty charts by effectively using critical contextual elements and showcasing adaptability to diverse chart types. These results highlight the necessity of contextual reasoning for high-performance chart understanding.

\section{Error Analysis}
\label{appendixa.2}
\autoref{fig:error_analysis} illustrates specific examples of the four major error types observed in the ChartMind: value recognition errors, judgment errors, calculation errors, and color recognition errors. These examples highlight typical failure cases, such as incorrect identification of numerical values in bar segments (value recognition), flawed logical reasoning or mismatched context interpretation (judgment), inaccurate arithmetic operations (calculation), and misassociation of chart elements with their respective colors in legends or overlapping areas (color recognition). The figure provides detailed scenarios, such as errors in identifying peak values, interpreting differences in chart segments, and miscalculating relationships between visual elements. These cases emphasize the challenges faced by models in aligning visual interpretation with reasoning accuracy.

\section{Potential Risks}

While our work primarily focuses on dataset construction and evaluation methodology for chart question answering (CQA), we acknowledge the following limited potential risks:

\begin{itemize}
    \item \textbf{Use of LLMs in Data Generation.} The initial QA pairs in ChartMind were generated using GPT-4o. Although all outputs were manually reviewed, revised, and filtered by trained annotators, there remains a low-level risk of inherited model bias or hallucination that may not have been fully eliminated.
    
    \item \textbf{Automated Evaluation.} Our experiments rely partially on GPT-4o for scoring open-ended answers. While we provide correlation analysis with human judgments to validate reliability (Section~5.2), model-based scoring may still carry implicit biases toward certain linguistic styles or answer formats.
    
    \item \textbf{Language Scope.} ChartMind currently supports English and Chinese. Although this already expands the field beyond English-only benchmarks, performance and fairness in other language contexts are not yet covered.
\end{itemize}

\noindent Overall, our design minimizes these risks through manual validation, diverse model comparisons, and detailed performance analysis. Future versions of ChartMind will incorporate broader language coverage and alternative evaluation strategies to further mitigate these concerns.

\end{document}